%% file: acl_latex.tex
\documentclass[11pt]{article}

\usepackage[final]{acl}

\usepackage{times}
\usepackage{latexsym}

\usepackage[T1]{fontenc}

\usepackage[utf8]{inputenc}

\usepackage{microtype}

\usepackage{inconsolata}

\usepackage{graphicx}

\usepackage[utf8]{inputenc} 
\usepackage[T1]{fontenc}    
\usepackage{hyperref}       
\usepackage{url}            
\usepackage{booktabs}       
\usepackage{amsfonts}       
\usepackage{nicefrac}       
\usepackage{microtype}      
\usepackage{xcolor}         
\usepackage[dvipsnames]{xcolor}

\usepackage{xspace}         
\usepackage{makecell}       
\usepackage{multirow}       
\usepackage{tablefootnote}
\usepackage{tabularx}  
\usepackage{array}     
\usepackage{booktabs}  
\usepackage{amsmath}   
\usepackage{longtable}
\usepackage{listings}

\usepackage{graphicx}
\usepackage{wrapfig}
\usepackage{subcaption}

\usepackage{enumitem}

\usepackage{algorithm}
\usepackage{algpseudocode}

\newcommand{\modelname}{{\textsc{FARON}}\xspace}

\title{Spatial Query Execution as Verifiable Rewards for Multi-step Spatial Reasoning}

\author{
 \textbf{Jiyoon Pyo\textsuperscript{1}},
 \textbf{Dongwon Jung\textsuperscript{2}},
 \textbf{Zekun Li\textsuperscript{1}},
 \textbf{Nayeon Kim\textsuperscript{1}},
\\
 \textbf{Muhao Chen\textsuperscript{2}},
 \textbf{Yao-Yi Chiang\textsuperscript{1}},
\\
 \textsuperscript{1}University of Minnesota - Twin Cities,
 \textsuperscript{2}University of California, Davis,
}

\begin{document}
\maketitle

\input{sections/0_abstract}
\input{sections/1_intro}
\input{sections/3_method}
\input{sections/4_experiment}

\input{sections/5_results}
\input{sections/6_related}
\input{sections/7_conclusion}

\section*{Limitations}
Our experiments rely on simplified synthetic scenes, consisting of randomly generated points, lines, polygons, and geometric objects (i.e., cubes). These controlled environments enable precise verification of spatial relations and intermediate reasoning states, yet they do not fully capture the visual diversity, object semantics, and scene complexity of real-world environments. However, we do not view this as a fundamental limitation of the \modelname pipeline. Because our framework operates on explicit object geometries and executable spatial relations, we can extend it to more realistic synthetic environments using assets such as Objaverse~\citep{deitke-2023-objaverse, deitke-2023-objaversexl} while retaining access to the geometric information required for verification. We leave the implementation and evaluation of such extensions to future work.

Our current 3D implementation also approximates egocentric reasoning through camera- and object-relative viewpoints rather than directly modeling embodied perspectives. While these settings capture certain viewpoint-dependent spatial relations, they do not reproduce the richer visual and contextual information present in embodied environments. Extending the framework to more realistic 3D scenes could enable broader evaluation of egocentric spatial reasoning.

Our experiments focus on improving spatial reasoning rather than visual perception. Our error analysis suggests that incorrect identification of intermediate object sets remains a source of failure, indicating that additional visual-grounding supervision may be beneficial. Furthermore, although we create canonical reasoning traces from verified spatial execution results, we do not independently validate the semantic fidelity of the Qwen3-32B paraphrases, which may introduce inconsistencies in the linguistic supervision. Nevertheless, our results show that this supervision improves spatial reasoning performance, particularly on multi-step questions.

\section*{Ethical Considerations}
This work follows the ACL Code of Ethics. The primary data used in this work are synthetically generated and do not involve human participants, personally identifiable information, or user-generated content. We use external benchmarks and pretrained models for research evaluation under their respective usage terms. Although \modelname targets multimodal spatial reasoning, the results should not be interpreted as validated for safety-critical applications such as autonomous navigation or robotics.



\bibliography{custom}

\appendix
\input{sections/A_appendix}
\end{document}

%% file: sections/0_abstract.tex
\begin{abstract}
Large vision-language models (LVLMs) remain unreliable on spatial reasoning problems that require composing grounded relations across multiple objects. Yet, scaling reasoning supervision is challenging since human-annotated visual datasets often rely on fixed relational annotations and coarse object localization, making it difficult to obtain precise intermediate reasoning states across diverse spatial configurations and complexities. We introduce \modelname (\textbf{F}ully \textbf{A}utomated \textbf{R}easoning \textbf{O}ver spatial relatio\textbf{N}s), a framework that executes spatial queries to provide both execution-grounded reasoning supervision and verifiable outcome rewards. Rather than relying on existing annotated data, \modelname constructs synthetic 2D and 3D scenes with explicit geometries, enabling configurable object arrangements, spatial relations, and multi-step compositions. Deterministic query execution yields intermediate sets and final answers, providing precise supervision without manual relation annotation. Our pipeline converts intermediate states into natural-language reasoning traces for supervised fine-tuning (SFT), while the execution-derived answers provide verifiable rewards for reinforcement learning with verifiable rewards (RLVR). Controlled experiments show that trace-based SFT improves multi-step spatial reasoning over final-answer supervision and that subsequent group relative policy optimization (GRPO) provides further gains. The resulting models also demonstrate transfer across external spatial reasoning benchmarks and multiple LVLM backbones without benchmark-specific fine-tuning. These results show the potential of using spatial query execution as a unified, flexible source of reasoning supervision and verifiable outcome rewards for multimodal spatial reasoning.
\end{abstract}

%% file: sections/1_intro.tex
\section{Introduction}
\begin{figure*}
    \centering
    \includegraphics[width=\linewidth]{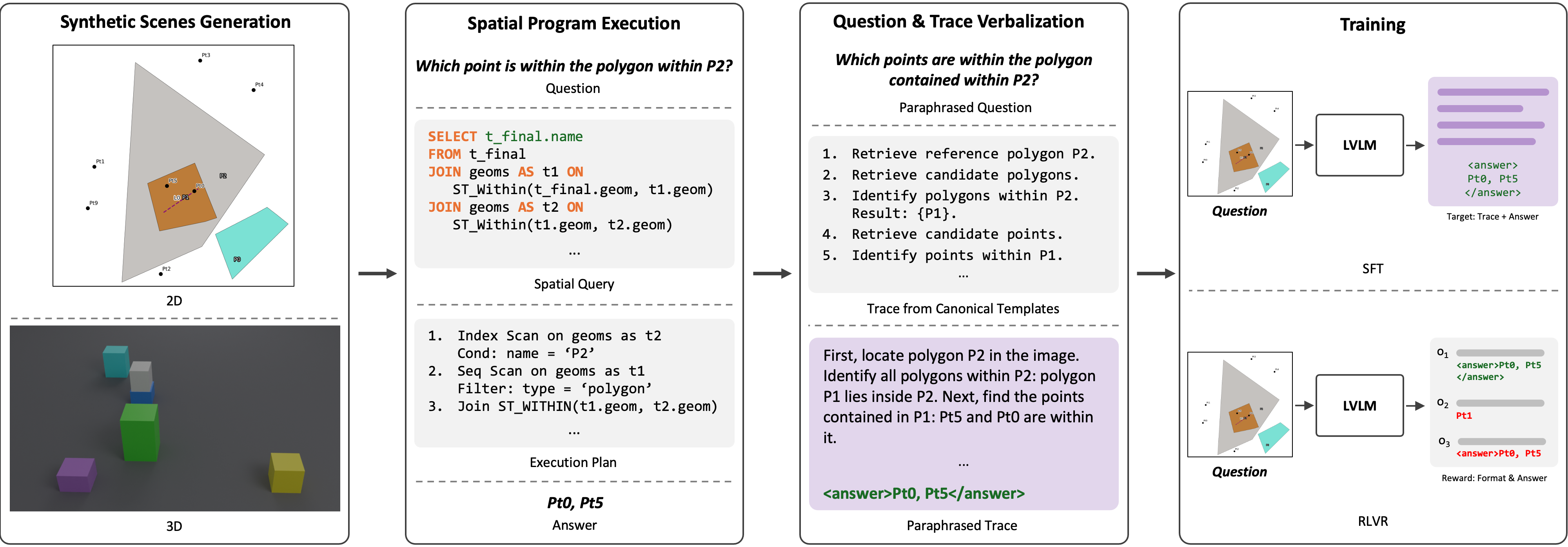}
    \caption{Overview of the FARON pipeline, from synthetic scene generation and spatial query execution to reasoning-trace supervision and verifiable reward training.}
    \label{fig:overall}
\end{figure*}


While large vision-language models (LVLMs) have made substantial progress on recognition-centric visual tasks, they remain unreliable when answering questions that require composing multiple grounded spatial relations~\citep{johnson-2017-clevr, liu-2023-vsr, kamath-2023-whatsup, stogiannidis-2025-mindthegap}. Answering questions involving containment, adjacency, intersection, relative direction, depth ordering, or occlusion requires more than recognizing which objects are present. A model must identify the relevant entities, determine their spatial relations, maintain intermediate constraints, and compose these results into a final answer.

One approach is to provide stronger supervision than final-answer examples, yet constructing reliable visual reasoning supervision remains challenging. Existing approaches~\citep{li-2025-spatialladder, batra-2025-spatialthinker, cai-2025-sensenovasi} often rely on human-written explanations~\citep{selvaraju-2019-takingahint, schwenk-2022-okvqa} or relational annotations such as scene graphs~\citep{krishna-2017-visualgenome, liu-2023-vsr}, which are costly to obtain at scale. For example, collecting the dense scene-graph annotations in Visual Genome required more than 33K unique workers over 21 months~\citep{krishna-2017-visualgenome}. Beyond annotation costs, relying on existing images limits control over object arrangement, the occurrence of certain spatial relations, and their compositional complexity. For this reason, prior work~\citep{johnson-2017-clevr} explored synthetic scene generation to control visual configurations and deliberately construct reasoning chains, rather than being restricted to naturally occurring configurations.

However, creating controlled visual scenes does not guarantee reliable supervision for intermediate reasoning. In spatial reasoning, each step requires identifying objects that satisfy specific geometric relations, and the correctness of these relations may depend on precise object boundaries and spatial configuration. Multi-step questions further introduce dependencies between intermediate object sets, where an incorrect result can propagate through subsequent spatial operations. Consequently, verifying manually constructed reasoning steps becomes increasingly difficult as more relations are composed, and incorrect rationales can provide misleading training signals~\citep{zhou-2024-noisyrationales}. Reinforcement learning with verifiable rewards (RLVR)~\citep{lambert-2024-tulu, shao-2024-grpo} offers a complementary training approach but requires outcomes that can be checked automatically and reliably~\citep{lambert-2024-tulu, shao-2024-grpo, li-2025-spatialladder, batra-2025-spatialthinker}. In other words, both approaches can benefit from a common, scalable source of precisely grounded supervision. This motivates our central question: \textit{Can we automatically generate both execution-grounded reasoning traces and verifiable outcome rewards for multimodal spatial reasoning at scale through a unified framework?}

We introduce \modelname (\textbf{F}ully \textbf{A}utomated \textbf{R}easoning \textbf{O}ver spatial relatio\textbf{N}s), a framework that constructs execution-grounded process supervision and verifiable rewards through synthetic scene generation and spatial query execution. Rather than relying on spatial annotations from existing images, \modelname generates two complementary forms of scenes with explicit object geometry. The 2D map-like scenes contain labeled points, lines, and polygons to support topological and directional reasoning, while the 3D rendered scenes contain geometric objects to support viewpoint-dependent direction, depth, and visibility reasoning. These environments capture different spatial reasoning requirements, allowing us to examine whether a common execution framework can support both boundary-dependent and viewpoint-dependent relations. Building on executable reasoning~\citep{chen-2023-pot, gao-2023-pal, jung-2025-code}, \modelname constructs spatial questions by composing executable relation blocks over the recorded scene geometries. Each block deterministically evaluates a spatial relation and produces verified intermediate sets, which can be passed to subsequent operations or combined with other intermediate sets to derive the final answer. This allows \modelname to automatically generate questions, intermediate reasoning states, and final answers through a unified execution framework for both 2D and 3D scenes. We convert these executed states into natural-language reasoning traces for supervised fine-tuning (SFT), while the execution-derived final answers provide automatically verifiable rewards for RLVR.

We first conduct internal evaluations to assess the effectiveness of \modelname-generated supervision, the integrity of its execution-grounded reasoning traces, and the feasibility of using execution-derived answers as verifiable rewards for RLVR through group relative policy optimization (GRPO). We then evaluate the resulting models on external spatial reasoning benchmarks to assess transfer across different visual settings, reasoning tasks, and LVLM backbones. These experiments show that execution-grounded supervision can improve multi-step spatial reasoning and transfer across diverse visual settings and LVLM backbones.

%% file: sections/3_method.tex
\section{Method}
\modelname generates spatial reasoning supervision through synthetic scene generation (Section~\ref{sec:synthetic-scene-generation}), spatial question construction and execution (Section~\ref{sec:reasoning-step-generation}), and natural-language trace generation (Section~\ref{sec:question-diversification}). The resulting questions, reasoning trace, and execution-derived answers support SFT and RLVR (Section~\ref{sec:training}). Figure~\ref{fig:overall} summarizes the pipeline, Appendix~\ref{app:algorithm} shows the pseudocode of our pipeline, and Appendix~\ref{app:e2e-sample} provides an end-to-end example. 


Each scene consists of a rendered image and a structured scene state containing object identifiers, exact geometric boundaries, spatial attributes, and, for 3D scenes, camera parameters. We run spatial queries on the recorded information to derive supervision targets, but we do not provide them to the LVLM during training or inference. 

\subsection{Synthetic Scene Generation} \label{sec:synthetic-scene-generation}
Relying on existing images limits both the control over the spatial configurations represented in the data and the precision of spatial supervision. Relying on available natural images constrains which object combinations and spatial relations can be sampled from the scene, limiting flexibility in constructing diverse spatial arrangements and reasoning compositions. In addition, human-annotated visual datasets commonly use bounding boxes for object localization, which do not preserve the exact geometry needed to reliably evaluate containment, intersection, and adjacency (Appendix~\ref{app:bbox-vs-boundary}). To address this limitation, \modelname generates synthetic scenes with configurable object arrangements and spatial relations, providing greater control over the spatial configurations used for question generation. By rendering these scenes from explicit object geometry, \modelname can also obtain precise geometric states for deterministic relation evaluation. We implement the framework for both 2D map-like and 3D rendered environments.

\paragraph{2D map-like scenes.}
Maps commonly represent spatial entities as points, lines, and polygons, with topological and directional relations describing their spatial arrangements. To capture these characteristics, we generate abstract 2D map-like scenes following defined sets of spatial predicates~\citep{clementini-1993-small}. We randomize object positions, sizes, colors, and labels to construct diverse configurations. Although the scenes simplify map visuals, they retain the geometric relationships required for spatial reasoning, such as containment and intersection. We compute spatial predicates using exact entity boundaries rather than approximate bounding boxes, enabling the generation of geometrically precise questions under controlled conditions. We show an example in Appendix~\ref{app:synthetic-data-example-2d}.

\paragraph{3D rendered scenes.}
We generate 3D scenes containing simple geometric objects to capture viewpoint-dependent spatial reasoning in multi-object environments. These scenes represent directional, depth, and visibility-based relations, such as \textit{left of} and \textit{behind}. Unlike 2D scenes, relations defined in the underlying 3D geometry may not be identifiable from rendered images due to occlusion and perspective projection. To ensure that questions are answerable from visual evidence rather than hidden geometric information, we restrict question generation to objects and spatial relations that are visually distinguishable in the rendered image. Therefore, we apply deterministic filters using the recorded geometry and camera parameters. We require objects to occupy a sufficiently large visible area and to retain a sufficient fraction of their projected area. We also require a minimum separation between projected object boundaries to reduce visual ambiguity. Only objects and relations satisfying these criteria are used for question generation. We provide the filtering thresholds and an example generated 3D scene in Appendix~\ref {app:method_filtering} and Appendix~\ref{app:synthetic-data-example-3d}, respectively.

\subsection{Question and Reasoning Trace Generation} \label{sec:reasoning-step-generation}
Given a synthetic scene with explicit geometry, \modelname constructs spatial questions by composing executable relation blocks. Each block specifies a spatial relation (e.g., left of, within, or intersects) between candidate and reference objects, which \modelname evaluates through a spatial query to return the subset of objects satisfying the relation. We provide the supported spatial predicates and their definitions in Appendix~\ref{app:method-relations}. While the special predicates and their terminology may differ across environments (e.g., \textit{east of} in 2D maps and \textit{right of} in 3D views), both follow the same execution framework for evaluating relations, recording intermediate sets, and composing subsequent operations. This allows \modelname to generate reasoning traces across different spatial relation families without separate trace-generation pipelines for 2D and 3D environments.

To provide supervision for both individual spatial relations and their compositions, we generate 1-step questions using a single relation block and multi-step questions by composing multiple blocks. Following the query construction strategy of \citet{huang-2025-targa}, we support chain-like and branching compositions, with up to five relation blocks in the reported experiments. In a chain, the intermediate set returned by a relation block becomes the input to the subsequent block. In branching compositions, we combine intermediate sets from different spatial constraints using operations such as intersection and difference. By varying the spatial predicates and dependencies between relation blocks, we generate questions with different relation types, reasoning depths, and compositional structures.

For each composed spatial query, we obtain its SQL execution plan to identify the underlying operations and execute the corresponding spatial subqueries to retrieve verified intermediate object sets. For example, consider the 2-step question \textit{``Which points are within the polygon contained within P2?''} (Figure~\ref{fig:overall}). The first spatial query identifies polygons contained within P2, yielding the intermediate set \textit{P1}. The second query identifies points within P1, yielding the final answer \textit{Pt0, Pt5}. We use these executed states to construct natural-language reasoning traces and final answers.  We use these operations and their verified results to construct execution-grounded reasoning traces. Appendix~\ref{app:example-executionplan} provides an example of the SQL execution plan and its corresponding intermediate states.

\paragraph{Filtering trivial compositions.} \label{sec:filtering-trivial-compositions}
We filter out multi-step questions in which combining spatial relations does not meaningfully affect the final answer. Specifically, we require every intermediate set to be non-empty and distinct across different branches. For intersection, the final result must be non-empty and different from both intermediate sets. For set difference, the final result must differ from the first intermediate set. These constraints prevent redundant spatial operations and encourage questions that require combining information from multiple reasoning steps.

\subsection{Natural-Language Trace Generation} \label{sec:question-diversification}
The execution-derived record contains spatial operations, verified intermediate object sets, and the final answer. We retain operations relevant to spatial reasoning while omitting database-specific operations, such as sorting and duplicate removal. We then use canonical templates to convert the resulting spatial operations and their intermediate states into natural-language reasoning steps. Each template specifies how to verbalize the spatial relation, its reference objects, and the resulting object set. By populating these templates directly from executed spatial query results, \modelname produces one natural-language reasoning step per relation block while preserving executed intermediate results. 
We provide canonical template definitions and examples in Appendix~\ref{app:method-relations}. 

To increase linguistic variability, we use Qwen3-32B~\citep{yang-2025-qwen3} to paraphrase the canonical questions and reasoning traces. We instruct the paraphrasing model to preserve object references, spatial relations, intermediate results, and answer semantics while varying only the linguistic form. We provide the paraphrasing prompts and examples in Appendix~\ref{app:method_paraphrasing} and Appendix~\ref{app:example-paraphrase}, respectively.

\subsection{Training with Reasoning Traces and Verifiable Rewards}\label{sec:training}
We use execution-derived intermediate states and final answers for two complementary training stages: process supervision and RLVR.

\paragraph{Process-supervised fine-tuning.}
Each training example contains a rendered image $I$, a question $q$, an execution-grounded reasoning trace $\tau$, and a final answer $a$. Given the image and question, the LVLM is trained to generate the target response $y$, consisting of the reasoning trace followed by the final answer enclosed in \texttt{<answer>} tags. We train to minimize the token-level cross-entropy loss:
\begin{equation}
    L_{SFT}=-\sum_{j=1}^{|y|}\log P(y_{j} \mid I, q, y_{<j})
\end{equation}
where $y_{j}$ is the target token at position $j$ and $y_{<j}$ denotes the preceding tokens. We use the underlying geometry and structured spatial query execution record to construct supervision targets but do not provide them to the model.

\paragraph{Reinforcement learning with verifiable rewards.}
We apply GRPO~\citep{shao-2024-grpo} to a disjoint set of \modelname-generated examples after trace-based SFT. For each image question pair $(I, q)$, the LVLM policy $\pi_\theta$ generates a group of $G$ responses $\{o_i\}_{i=1}^{G}$. Each response receives a reward based on its final-answer correctness and output-format compliance:

\begin{equation}
    r_i = R_{answer}(o_i,a) 
    + \lambda R_{format}(o_i)
\end{equation}
where $a$ is the execution-derived ground-truth answer. $R_{answer}(o_i,a)$ equals 1 when the extracted answer matches $a$, and $R_{format}(o_i)$ equals 1 when the response follows the required output format. GRPO normalizes rewards within each sampled group to compute group-relative advantages $A_{i}$:
\begin{equation}
    \hat{A}_i =
    \frac{r_i-\bar{r}}{\sigma_r+\delta},
\end{equation}
where $r_i$ is the reward assigned to the $i$-th sampled response, $\bar{r}$ and $\sigma_r$ are the mean and standard deviation of the group rewards, respectively, and $\delta$ is a small constant added for numerical stability. We use these advantages to optimize the LVLM policy.

Since the final answer is obtained through deterministic spatial execution, we can score model responses without human annotation or a learned reward model. 
We provide additional optimization details in Table~\ref{tab:grpo-hparams} in Appendix~\ref{app:training-hyperparameters}.

%% file: sections/4_experiment.tex
\section{Experimental Setting} \label{sec:experimental-setting}
We organize our experiments into internal and external evaluations. Internal evaluation (RQ1-RQ3) uses held-out \modelname-generated questions to assess the effectiveness of the generated supervision and its use in training, while external evaluation (RQ4-RQ5) uses spatial reasoning benchmarks to test transfer beyond the synthetic training distribution and across LVLM backbones. We test the following research questions: 

\noindent \textbf{RQ1}: Does execution-grounded reasoning-trace supervision improve spatial reasoning beyond answer-only SFT?

\noindent \textbf{RQ2}: How do modifications to intermediate information, execution order, and state correctness affect the benefits of reasoning-trace supervision?

\noindent \textbf{RQ3}: Can \modelname-generated examples support RLVR, and does GRPO provide additional gains after trace-based SFT?

\noindent \textbf{RQ4}: Does the trace-based SFT+GRPO pipeline transfer to external spatial reasoning benchmarks without benchmark-specific fine-tuning?

\noindent \textbf{RQ5}: Are the observed transfer gains consistent across LVLM families and parameter scales?
    
\paragraph{Training data and splits.}
The SFT data contain 23,800 questions: 11,590 generated from 2,199 2D scenes and 12,210 from 1,256 3D scenes. The GRPO data contain 1,736 questions generated from a separate set of 289 2D and 186 3D scenes. Internal evaluation uses 8,445 questions from another held-out set of 354 2D and 122 3D scenes. All splits are constructed at the scene level so that the underlying scenes are disjoint across SFT, GRPO, and internal evaluation. 

\paragraph{Training variants.}
For RQ1-RQ3, we evaluate seven variants using Qwen2.5-VL-3B as the LVLM backbone. \textbf{Base} denotes the original instruction-tuned checkpoint. \textbf{\modelname-Answer} uses answer-only SFT, while \textbf{\modelname-Reason} uses execution-derived reasoning traces followed by final answers. To assess trace integrity, \textbf{\modelname-Mask} removes selected intermediate results, \textbf{\modelname-Shuffle} reorders multi-step reasoning traces, and \textbf{\modelname-Corrupt} introduces incorrect intermediate predicates or object sets. \textbf{\modelname-Reason+GRPO} applies GRPO after trace-based SFT using execution-derived final answers as verifiable rewards. All controlled SFT variants use identical training examples, data ordering, optimization settings, and evaluation procedures, differing only in their supervision targets.

\paragraph{External benchmarks and baselines.}
To address RQ4-RQ5, we evaluate transfer to five external spatial reasoning benchmarks with varying reasoning requirements. CV-Bench~\citep{tong-2024-cambrian} and BLINK~\citep{fu-2024-blink} primarily assess individual spatial and perceptual judgments, including object counting, spatial relations, relative depth, and distance. SpatialEval~\citep{wang-2024-spatialeval} covers diverse spatial tasks, including relation understanding and navigation, while FRIEDA~\citep{pyo-2026-frieda} explicitly requires multi-step 2D-map-based reasoning. Multihop-Spatial~\citep{lee-2026-multihopspatial} evaluates spatial compositions spanning one to three reasoning hops. These benchmarks allow us to examine transfer to both basic spatial capabilities and multi-step spatial reasoning. None of these benchmarks are used during \modelname's SFT or GRPO training. For comparison, we locally evaluate publicly available checkpoints of spatially specialized baselines~\citep{li-2025-spatialladder, batra-2025-spatialthinker, cai-2025-sensenovasi, wu-2026-vilasr, lee-2026-multihopspatial}.

\paragraph{Models.}
To address RQ5, we evaluate \modelname across Qwen2.5-VL-3B, Qwen2.5-VL-7B~\citep{bai-2025-qwen25vl}, Qwen3-VL-4B~\citep{bai-2025-qwen3vl}, and InternVL3-2B~\citep{zhu-2025-internvl3}. This selection covers variation in model scale, version, and architecture. We use the same training hyperparameter configuration across backbones. We report full training configurations, evaluation settings, and computational resources in Appendix~\ref{app:training-hyperparameters}.

%% file: sections/5_results.tex
\section{Experiment Results} \label{sec:experiment-results}
We organize the results according to the five research questions introduced in Section~\ref{sec:experimental-setting}. 


\subsection{Effect of Intermediate Supervision} \label{sec:result-intermediate}
We compare \modelname-Answer and \modelname-Reason on held-out \modelname scenes to assess whether execution-grounded reasoning trace supervision improves spatial reasoning over answer-only supervision. 

\begin{table}[h!]
    \centering
    \small
    \begin{tabular}{lccc}
    \toprule
        Model Variant
        & \textbf{Overall}
        & \textbf{1-step} 
        & \textbf{5-step} \\
    \midrule
        Base & 12.97 & 18.80 & 5.83 \\
        \modelname-Answer & 59.94 & \textbf{80.63} & 44.74 \\
        \modelname-Reason & \textbf{67.53} & 78.60 & \textbf{55.08} \\
    \bottomrule
    \end{tabular}
    \caption{Accuracy (\%) on held-out \modelname scenes for the base model, answer-only SFT, and trace-supervised SFT.}
    \label{tab:internal_sft_results}
\end{table}

\paragraph{Analysis.}
Both \modelname-Answer and \modelname-Reason substantially outperform the base model, demonstrating the effectiveness of \modelname-generated data for spatial reasoning. Moreover, \modelname-Reason outperforms \modelname-Answer by 7.59pp overall, with a larger gain on 5-step questions (+10.34pp). In contrast, \modelname-Answer slightly outperforms \modelname-Reason on 1-step questions, where intermediate spatial composition is not required. These results highlight the additional benefits of \modelname's execution-grounded reasoning supervision beyond answer-only SFT, particularly for multi-step spatial reasoning.

\subsection{Effect of Trace Integrity} \label{sec:result-trace-integrity}
We compare \modelname-Reason against the Mask, Shuffle, and Corrupt variants to assess the importance of preserving intermediate reasoning information.  

\begin{table}[h!]
    \centering
    \small
    \begin{tabular}{lccc}
    \toprule
        Model Variant
        & \textbf{Overall}
        & \textbf{1-step} 
        & \textbf{5-step} \\
    \midrule
        \modelname-Reason & \textbf{67.53} & \textbf{78.60} & \textbf{55.08} \\
        \modelname-Mask & 65.65 & 78.47 & 52.30 \\
        \modelname-Shuffle & 64.38 & 78.03 & 50.62 \\
        \modelname-Corrupt & 61.97 & 76.63 & 40.42 \\
    \bottomrule
    \end{tabular}
    \caption{Accuracy (\%) on held-out \modelname scenes under controlled reasoning-trace modifications.}
    \label{tab:trace_integrity}
\end{table}

\paragraph{Analysis.}
\modelname-Reason achieves the highest accuracy across all settings, with pronounced gains on 5-step questions. \modelname-Corrupt shows the largest degradation (-14.66pp), highlighting the value of accurate verified intermediate states for compositional spatial reasoning. 
Differences on 1-step questions remain relatively small. 
These results indicate that preserving intermediate states and their dependencies is particularly important for multi-step spatial reasoning.

\subsection{Effect of RLVR with GRPO} \label{sec:result-grpo}
We compare \modelname-Reason with \modelname-Reason+GRPO on held-out \modelname scenes to assess whether RLVR provides additional gains after trace-based SFT.

\begin{table}[h!]
    \centering
    \small
    \begin{tabular}{lccc}
    \toprule
        Model Variant
        & \textbf{Overall}
        & \textbf{1-step} 
        & \textbf{5-step} \\
    \midrule
        \modelname-Reason & 67.53 & 78.60 & 55.08 \\
        \modelname-Reason+GRPO & \textbf{70.46} & \textbf{78.63} & \textbf{59.14} \\
    \bottomrule
    \end{tabular}
    \caption{Accuracy (\%) on held-out \modelname scenes before and after GRPO using execution-derived outcome rewards.}
    \label{tab:grpo_results}
\end{table}

\paragraph{Analysis.}
GRPO using \modelname-generated verifiable rewards further improves accuracy over trace-based SFT, particularly on 5-step questions (+4.06pp), while 1-step performance remains nearly unchanged. These results demonstrate that \modelname supports both execution-grounded process supervision and automatically verifiable outcome rewards for RLVR through a unified generation framework.

\subsection{External Transfer Without Benchmark-Specific Fine-Tuning} \label{sec:result-transfer}

We evaluate \modelname-Reason+GRPO on five external spatial reasoning benchmarks not used during training. Table~\ref{tab:external_overall_results} compares its overall performance with corresponding base checkpoints and spatially specialized models, with detailed subcategory results provided in Appendix~\ref{app:detailed-results}.

\begin{table*}[h!]
    \centering
    \small
    \begin{tabular}{lccccc}
    \toprule
        \multirow{2}{*}{Model Variant}
        & \multicolumn{2}{c}{\textbf{2D Spatial Reasoning}}
        & \multicolumn{3}{c}{\textbf{3D Spatial Reasoning}} \\
        \cmidrule(lr){2-3} \cmidrule(lr){4-6}
        & SpatialEval
        & FRIEDA
        & CV-Bench
        & BLINK\footnotemark[1]
        & MultiHopSpatial \\
        \midrule
        \textbf{Qwen2.5-VL-3B-based models} & & & & & \\
        Base                    & 56.76 & 16.40 & 71.24 & 78.01 & 33.87 \\
        SpatialLadder           & 52.20 & 15.80 & 75.87 & 77.12 & 28.18 \\
        SpatialThinker          & 56.33 & 16.80 & 73.42 & 78.97 & \underline{34.58} \\
        SenseNova-SI-1.1        & \underline{57.98} & \underline{18.80} & \underline{76.65} & \underline{83.09} & 33.64 \\
        \modelname-Reason+GRPO  & \color{Green}\textbf{60.24} & \textbf{19.80} & \color{Green}\textbf{80.16} & \color{Green}\textbf{84.35} & \color{Green}\textbf{39.87} \\

        \midrule
        \textbf{Qwen2.5-VL-7B-based models} & & & & & \\
        Base                    & 57.07 & 18.20 & 78.50 & 84.67 & 34.24 \\
        SpatialThinker          & 61.00 & 19.40 & 79.91 & \underline{85.37} & \underline{34.38} \\
        SenseNova-SI-1.1        & \textbf{70.24} & \underline{20.40} & \underline{79.92} & \textbf{87.95} & 32.33 \\
        ViLaSR                  & 61.38 & 14.80 & 75.18 & 85.08 & 28.47 \\
        \modelname-Reason+GRPO  & \underline{67.06} & \textbf{21.60} & \color{Green}\textbf{81.86} & \color{Mulberry}84.38 & \color{Green}\textbf{38.31} \\

        \midrule
        \textbf{Qwen3-VL-4B-based models} & & & & & \\
        Base                    & \underline{57.88} & 12.80 & \textbf{83.59} & \underline{84.08} & \underline{37.04} \\
        MultihopSpatial         & 56.84 & \underline{13.60} & 81.94 & 82.68 & --\footnotemark[2] \\ 
        \modelname-Reason+GRPO  & \color{Green}\textbf{62.15} & \color{Green}\textbf{20.80} & \color{Mulberry}\underline{83.55} & \color{Green}\textbf{87.25} & \color{Green}\textbf{40.11} \\

        \midrule
        \textbf{InternVL3-2B-based models} & & & & & \\
        Base                    & 24.79 & \textbf{16.60} & 74.41 & \underline{77.57} & 34.93 \\
        SenseNova-SI-1.1        & \textbf{53.41} & \underline{14.80} & \underline{77.39} & 73.86 & \textbf{39.82} \\
        \modelname-Reason+GRPO  & \underline{49.13} & \color{Mulberry}\underline{14.80} & \color{Green}\textbf{79.20} & \textbf{78.84} & \color{Mulberry}\underline{39.78} \\

    \bottomrule
    \end{tabular}

    \caption{Overall scores on external spatial reasoning benchmarks. \modelname-Reason+GRPO is evaluated without benchmark-specific fine-tuning. \textbf{Bold} and \underline{underlined} values indicate the highest and second-highest scores within each backbone group and benchmark. \color{Green}\textbf{Green} \color{black}indicates statistically significant improvement ($p<0.05$) over the strongest competing model, while \color{Mulberry}purple \color{black}indicates a lower but statistically insignificant score ($p\geq0.05$) relative to the strongest competing model. We assess statistical significance using 10,000 paired bootstrap resamples.}
    \label{tab:external_overall_results}
\end{table*}


\paragraph{Analysis.} 
\modelname-Reason+GRPO outperforms the corresponding base checkpoints on SpatialEval and MultiHopSpatial across all four backbones. On SpatialEval, we observe the largest improvement for InternVL3-2B (+24.34pp). On FRIEDA, three of the four backbones improve over their base checkpoints. Results on CV-Bench and BLINK vary by backbone, with Qwen2.5-VL-3B achieving the highest scores among its compared models across all five benchmarks. These results indicate that supervision generated from synthetic spatial scenes can transfer to both map-based and image-based reasoning tasks.

\footnotetext[1]{The overall BLINK score is computed by macro-averaging the Spatial Relation and Depth subset scores.}
\footnotetext[2]{The MultiHopSpatial baseline achieves 62.40\% on MultiHopSpatial, but this score is omitted from the table as the model was trained on the benchmark's data and thus does not represent external transfer.}

\subsection{Consistency Across Model Families and Scales} \label{sec:result-backbones}
We examine whether \modelname's benefits generalize across LVLM backbones differing in parameter scale (Qwen2.5-VL-3B and Qwen2.5-VL-7B), model version (Qwen2.5-VL and Qwen3-VL), and architecture (QwenVL and InternVL). 

\paragraph{Analysis.}
\modelname-Reason+GRPO improves over the corresponding base checkpoint in 17 of 20 backbone-benchmark comparisons. These results suggest that execution-grounded supervision transfers across LVLM families and scales, suggesting its benefits are not limited to a single backbone.

\color{black}

%% file: sections/6_related.tex
\section{Related Work} \label{sec:related-work}

\paragraph{Spatial reasoning in LVLMs.}
Spatial reasoning remains a persistent weakness of LVLMs across both abstract and natural visual settings. Prior work evaluates reasoning over synthetic scenes with explicitly defined object relations~\citep{johnson-2017-clevr}, relative spatial relations in natural images~\citep{liu-2023-vsr}, and broader capabilities such as 2D counting, spatial relations, depth, and distance reasoning~\citep{tong-2024-cambrian, fu-2024-blink}. Other studies extend spatial evaluation to map-, grid-, and maze-based settings involving positional understanding, counting, and navigation~\citep{wang-2024-spatialeval}, as well as multi-step reasoning over maps extracted from reports~\citep{pyo-2026-frieda}. Several recent benchmarks further require models to combine multiple spatial constraints across reasoning steps rather than resolve a single spatial relation~\citep{wang-2024-spatialeval, pyo-2026-frieda, lee-2026-multihopspatial}. These studies show that strong visual recognition performance does not necessarily translate to reliable grounded spatial reasoning, particularly when success depends on maintaining and composing intermediate spatial states.

\paragraph{Improving spatial reasoning in LVLMs.}
Recent work improves LVLM spatial reasoning through spatially grounded supervision and reasoning-oriented training. SpatialRGPT~\citep{cheng-2024-spatialrgpt} curates training data from 3D scene graphs, while SpatialVLM~\citep{chen-2024-spatialvlm} automatically generates spatial question-answer pairs using object detection, segmentation, and estimated depth from real images. SenseNova-SI~\citep{cai-2025-sensenovasi} scales spatial supervision through large-scale data curation, and SpatialLadder~\citep{li-2025-spatialladder} combines spatial annotations from reconstructed scenes with progressive training and RLVR. More recently, SpatialThinker~\citep{batra-2025-spatialthinker} uses dense spatial rewards, while ViLaSR~\citep{wu-2026-vilasr} introduces visual drawing-based reasoning supervision. These approaches demonstrate the benefits of spatially grounded supervision but rely on existing images or estimated geometry, limiting control over the spatial configurations and intermediate states used for training. In contrast, \modelname uses synthetic geometry to generate diverse spatial configurations, derive verifiable intermediate reasoning states, and obtain deterministic final answers. 

\paragraph{Programmatic and executable reasoning supervision.}
A closely related idea uses executable computation to ground intermediate reasoning rather than relying on free-form rationales. Program-of-Thoughts~\citep{chen-2023-pot} delegates intermediate computations to executable programs while PAL~\citep{gao-2023-pal} generates programs whose execution produces the final answer. Recently, \citet{jung-2025-code} uses intermediate program execution states as supervision by translating execution traces into natural-language reasoning steps. We extend this idea to multimodal spatial reasoning. Instead of grounding intermediate computation in variables and program states, \modelname grounds each step in objects and spatial relations from an explicit geometric scene representation. Our pipeline derives the questions, intermediate object sets, spatial operations, and final answers through the same execution process and verbalizes the resulting states as natural-language reasoning traces for LVLM training. In this way, \modelname adapts executable process supervision from symbolic and language reasoning to grounded visual reasoning, where intermediate states must remain consistent not only with the computation but also with the underlying visual scene. 

While prior visual reasoning work has used structured programs, execution typically performs the reasoning procedure, invokes external tools, or verifies generated data. IEP~\citep{johnson-2017-iep} represents visual questions as programs and executes the corresponding neural modules to predict the final answer. VPD~\citep{hu-2024-vpd} executes and verifies tool-using visual programs and distills programmatic behavior into a vision-language model, while SPRITE~\citep{helu-2025-sprite} uses executable programs and simulator metadata to construct and verify spatial question-answer data. \modelname differs in how execution supports training: spatial queries over explicit scene geometries produce verified intermediate object sets for natural-language reasoning supervision, while their final answers provide verifiable rewards for GRPO. The same spatial execution pipeline provides both intermediate reasoning supervision and outcome-level rewards.

%% file: sections/7_conclusion.tex
\section{Conclusion}
We introduce \modelname, a framework that uses spatial query execution to automatically generate both process supervision and verifiable outcome rewards for LVLM spatial reasoning. By representing synthetic 2D and 3D scenes with their underlying geometry, \modelname deterministically derives intermediate sets and final answers through a common execution process. We verbalize intermediate states into natural-language reasoning traces for SFT, while execution-derived answers provide automatically verifiable rewards for RLVR.

Our experiments demonstrate the benefits of \modelname across both training stages. First, execution-grounded process supervision improves spatial reasoning beyond answer-only SFT, particularly on multi-step questions. Controlled modifications to the reasoning traces reduce performance, demonstrating that the verified intermediate steps generated by \modelname provide effective supervision when verbalized into natural-language reasoning traces. Second, GRPO using execution-derived outcome rewards provides additional gains over trace-based SFT. The resulting models also demonstrate transfer to external 2D and 3D spatial reasoning benchmarks across different LVLM backbones without benchmark-specific fine-tuning. These results show the potential of spatial query execution for scalable, execution-grounded supervision of spatial reasoning.

%% file: sections/A_appendix.tex
\input{sections/appendix/add_method}

\input{sections/appendix/add_examples}
\input{sections/appendix/add_results}
\input{sections/appendix/add_licenses}

%% file: sections/appendix/add_method.tex
\section{Additional Method Details}

\subsection{Details on Data Generation} \label{app:algorithm}
Algorithm~\ref{alg:faron-generation} summarizes the procedure for generating execution-grounded spatial reasoning examples. The algorithm applies to both 2D map-like and 3D rendered scenes, with additional visibility and relation-level filtering for 3D scenes (Appendix~\ref{app:method_filtering}).

\begin{algorithm}[h]
\caption{\modelname example generation for one scene}
\label{alg:faron-generation}
\footnotesize
\begin{algorithmic}
\Require scene configuration $c$, maximum relation blocks $K=5$
\Ensure examples $\mathcal{D}$ of $(I,q,\tau,a)$

\State $\mathcal{D} \gets \emptyset$
\State $(I,G) \gets \Call{GenerateScene}{c}$
\State $V \gets \Call{FilterObjects}{I,G}$
\State $\mathcal{B} \gets \Call{SampleCompositions}{G,V,K}$

\ForAll{$B \in \mathcal{B}$}
    \If{$\neg \Call{ValidRelations}{B,I,G}$}
        \State \textbf{continue}
    \EndIf

    \State $Q \gets \Call{CompileSQL}{B}$
    \State $P \gets \Call{ExecutionPlan}{Q}$
    \State $O \gets \Call{CurateOperations}{P}$
    \State $S \gets \Call{ExecuteSubqueries}{B,G}$
    \If{$\neg \Call{ValidComposition}{B,S}$}
        \State \textbf{continue}
    \EndIf

    \State $a \gets \Call{ExecuteSQL}{Q,G}$
    \State $R \gets \Call{GroundOperations}{O,B,S,a}$
    \State $(q_c,\tau_c) \gets \Call{Canonicalize}{B,R,a}$
    \State $(\mathcal{Q},\mathcal{T}) \gets \Call{Paraphrase}{q_c,\tau_c}$
    \State $\mathcal{D} \gets \mathcal{D} \cup$
    \Statex \hspace{\algorithmicindent}
    $\Call{ConstructExamples}{I,q_c,\tau_c,\mathcal{Q},\mathcal{T},a}$
\EndFor

\State \Return $\mathcal{D}$
\end{algorithmic}
\end{algorithm}

\subsection{Spatial Relation Definitions and Trace Templates} \label{app:method-relations}
We define spatial predicates between a subject
object $A$ and a reference object $B$. For 2D
scenes, topological relations are evaluated
directly on the recorded point, line, and polygon
geometries. Directional relations use coordinate
extrema derived from these geometries. For 3D
scenes, we use the projected visible boundary
polygons and recorded camera-frame depth. Table~\ref{tab:sql-predicates} and Table~\ref{tab:sql-predicates-3d} summarize the spatial predicates used for query execution, while Table~\ref{tab:trace-templates-relations} lists their canonical natural-language templates.

\paragraph{Notation and thresholds.}
Let $g_A$ and $g_B$ denote the exact geometries of objects $A$ and $B$, respectively, and $\partial g$ denote a geometry's boundary. For directional comparisons, we define $x_A^-$, $x_A^+$, $y_A^-$, and $y_A^+$ as the minimum and maximum coordinates of $A$, computed directly from its geometric boundary. The same notation applies to $B$. For 3D scenes, we derive these extrema from the projected visible boundary polygons.

Directional margins are defined as $m_x=0.005W$ and $m_y=0.005H$, where $W$ and $H$ denote the corresponding coordinate ranges. For 3D scenes, corner depths are computed as $z=-z_{\mathrm{camera}}$ for all eight corners of an object's 3D bounding volume. Depth increases with distance from the camera, and front/behind comparisons require a minimum separation of $\delta_{AB}=0.02\max(z_A,z_B)$.

\begin{table*}[t]
\centering
\footnotesize
\setlength{\tabcolsep}{5pt}
\renewcommand{\arraystretch}{1.08}

\begin{tabularx}{\textwidth}{
@{}p{0.20\textwidth}
>{\raggedright\arraybackslash}X@{}}
\toprule
\textbf{Relation} & \textbf{Spatial predicate or condition} \\
\midrule

\multicolumn{2}{@{}l}{\textbf{2D topological relations}} \\

within
& \texttt{ST\_Within}$(g_A,g_B)$ and
  $d(g_A,\partial g_B)\geq g$. \\

contains
& \texttt{ST\_Contains}$(g_A,g_B)$ and
  $d(g_B,\partial g_A)\geq g$. \\

covers
& \texttt{ST\_Covers}$(g_A,g_B)$. \\

disjoint
& \texttt{ST\_Disjoint}$(g_A,g_B)$ and
  $d(g_A,g_B)\geq g$. \\

partly overlapping
& Polygon--polygon
  \texttt{ST\_Overlaps}$(g_A,g_B)$,
  $\operatorname{area}(g_A\cap g_B)\geq6$,
  and $\operatorname{area}(g_A\setminus g_B)\geq6$. \\

touches
& \texttt{ST\_Touches}$(g_A,g_B)$. \\

shares border with
& \texttt{ST\_Touches}$(g_A,g_B)$ and
  $\operatorname{length}(\partial g_A
  \cap\partial g_B)\geq4$. \\

intersects
& \texttt{ST\_Intersects}$(g_A,g_B)$. \\

crosses
& \texttt{ST\_Crosses}$(g_A,g_B)$.
  For line--polygon crossing, at least $2g$
  of the line must lie inside and outside
  the polygon, respectively. \\

lies on
& \texttt{ST\_Covers}$(g_B,g_A)$,
  where $A$ is a point and $B$ is a line. \\

lies on border of
& For a point $A$,
  \texttt{ST\_Covers}$(\partial g_B,g_A)$.
  For a line $A$, at least $90\%$ of its
  length must lie within $0.05$ units of
  $\partial g_B$. \\

\midrule
\multicolumn{2}{@{}l}{\textbf{2D directional relations}} \\

east of
& $x_A^-\geq x_B^++m_x$. \\

west of
& $x_A^+\leq x_B^--m_x$. \\

north of
& $y_A^-\geq y_B^++m_y$. \\

south of
& $y_A^+\leq y_B^--m_y$. \\

northeast of
& East of and north of $B$. \\

northwest of
& West of and north of $B$. \\

southeast of
& East of and south of $B$. \\

southwest of
& West of and south of $B$. \\

\bottomrule
\end{tabularx}

\caption{Spatial predicates for 2D map-like
scenes. Topological relations are evaluated
directly on exact object geometries, while
directional relations compare coordinate
extrema derived from those geometries.}
\label{tab:sql-predicates}
\end{table*}

\begin{table*}[t]
\centering
\footnotesize
\setlength{\tabcolsep}{5pt}
\renewcommand{\arraystretch}{1.08}

\begin{tabularx}{\textwidth}{
@{}p{0.20\textwidth}
>{\raggedright\arraybackslash}X@{}}
\toprule
\textbf{Relation} & \textbf{Spatial predicate or condition} \\
\midrule

\multicolumn{2}{@{}l}{\textbf{3D projected-layout relations}} \\

right of
& $x_A^-\geq x_B^++m_x$,
  using extrema of the projected visible
  boundary polygons. \\

left of
& $x_A^+\leq x_B^--m_x$,
  using extrema of the projected visible
  boundary polygons. \\

above
& Vertical separation exceeding $m_y$,
  with horizontal support overlap,
  depth alignment, and boundary-band
  checks derived from projected visible
  polygon geometries. \\

below
& Symmetric to \textit{above}, with $A$
  below $B$ under the corresponding
  projected geometric and depth constraints. \\

\midrule
\multicolumn{2}{@{}l}{\textbf{3D depth relations}} \\

behind
& $z_A-z_B>\delta_{AB}$, where
  $\delta_{AB}=0.02\max(z_A,z_B)$. \\

in front of
& $z_B-z_A>\delta_{AB}$. \\

behind-right
& Behind and right of $B$. \\

behind-left
& Behind and left of $B$. \\

front-right
& In front of and right of $B$. \\

front-left
& In front of and left of $B$. \\

closer to camera
& $A.\mathrm{depth}_{\min}
  <B.\mathrm{depth}_{\min}$. \\

farther from camera
& $A.\mathrm{depth}_{\min}
  >B.\mathrm{depth}_{\min}$. \\

\midrule
\multicolumn{2}{@{}l}{\textbf{3D occlusion relations}} \\

partially occluding
& Projected geometries overlap, and $A$
  is closer than $B$ in camera-frame depth. \\

partially occluded by
& Projected geometries overlap, and $A$
  is farther than $B$ in camera-frame depth. \\

\midrule
\multicolumn{2}{@{}l}{\textbf{Comparative distance relations}} \\

closer to $X$
& For two candidates compared with reference
  $X$, let $d_{\mathrm{near}}$ and
  $d_{\mathrm{far}}$ denote their respective
  distances to $X$.
  In 3D, require
  $d_{\mathrm{far}}/d_{\mathrm{near}}\geq1.3$
  and $d_{\mathrm{far}}-d_{\mathrm{near}}\geq0.75$.
  In 2D, require a distance gap of at least $1.5$. \\

farther from $X$
& Uses the same comparison margins,
  selecting the candidate with the
  greater distance to $X$. \\

\bottomrule
\end{tabularx}

\caption{Projected-layout, depth, occlusion,
and comparative-distance predicates for
3D rendered scenes. Spatial relations
are computed using projected visible
boundary polygons and camera-frame depth.
Directional comparisons use extrema
derived from the projected boundaries,
with margins of $0.5\%$ of image width
or height. Front/behind relations require
a depth separation exceeding $2\%$ of
the larger compared depth. All relations
are evaluated after the visibility filters
in Appendix~\ref{app:method_filtering}.}
\label{tab:sql-predicates-3d}
\end{table*}

\paragraph{Canonical reasoning templates.} We construct canonical reasoning steps by
verbalizing executed spatial relations and their
verified intermediate object sets.
Table~\ref{tab:trace-templates-relations}
presents the templates for each supported relation.

\begin{table*}[tbph]
\centering
\footnotesize
\setlength{\tabcolsep}{5pt}
\renewcommand{\arraystretch}{1.05}

\begin{tabularx}{\textwidth}{
@{}p{0.21\textwidth}
>{\raggedright\arraybackslash}X@{}}
\toprule
\textbf{Relation} & \textbf{Canonical reasoning template} \\
\midrule

\multicolumn{2}{@{}l}{\textbf{2D topological relations}} \\

within
& The \{subject\} is within the \{reference\}. \\

contains
& The \{subject\} contains the \{reference\}. \\

covers
& The \{subject\} covers the \{reference\}. \\

disjoint
& The \{subject\} is disjoint from the \{reference\}. \\

partly overlapping
& The \{subject\} partly overlaps the \{reference\}. \\

touches
& The \{subject\} touches the \{reference\}. \\

shares border with
& The \{subject\} shares a border with the \{reference\}. \\

intersects
& The \{subject\} intersects the \{reference\}. \\

crosses
& The \{subject\} crosses the \{reference\}. \\

lies on
& The \{subject\} lies on the \{reference\}. \\

lies on border of
& The \{subject\} lies on the border of the \{reference\}. \\

\midrule
\multicolumn{2}{@{}l}{\textbf{2D directional relations}} \\

east of
& The \{subject\} is east of the \{reference\}. \\

west of
& The \{subject\} is west of the \{reference\}. \\

north of
& The \{subject\} is north of the \{reference\}. \\

south of
& The \{subject\} is south of the \{reference\}. \\

northeast of
& The \{subject\} is northeast of the \{reference\}. \\

northwest of
& The \{subject\} is northwest of the \{reference\}. \\

southeast of
& The \{subject\} is southeast of the \{reference\}. \\

southwest of
& The \{subject\} is southwest of the \{reference\}. \\

\midrule
\multicolumn{2}{@{}l}{\textbf{3D projected-layout relations}} \\

right of
& From the camera view, the \{subject\} is to
  the right of the \{reference\}. \\

left of
& From the camera view, the \{subject\} is to
  the left of the \{reference\}. \\

above
& From the camera view, the \{subject\} is
  above the \{reference\}. \\

below
& From the camera view, the \{subject\} is
  below the \{reference\}. \\

\midrule
\multicolumn{2}{@{}l}{\textbf{3D depth relations}} \\

behind
& The \{subject\} is behind the \{reference\}. \\

in front of
& The \{subject\} is in front of the \{reference\}. \\

behind-right
& The \{subject\} is behind and to the
  right of the \{reference\}. \\

behind-left
& The \{subject\} is behind and to the
  left of the \{reference\}. \\

front-right
& The \{subject\} is in front of and to
  the right of the \{reference\}. \\

front-left
& The \{subject\} is in front of and to
  the left of the \{reference\}. \\

closer to camera
& The \{subject\} is closer to the camera
  than the \{reference\}. \\

farther from camera
& The \{subject\} is farther from the camera
  than the \{reference\}. \\

\midrule
\multicolumn{2}{@{}l}{\textbf{3D occlusion relations}} \\

partially occluding
& The \{subject\} partially occludes
  the \{reference\}. \\

partially occluded by
& The \{subject\} is partially occluded
  by the \{reference\}. \\

\midrule
\multicolumn{2}{@{}l}{\textbf{Comparative distance relations}} \\

closer to $X$
& The \{subject\} is closer to the
  \{target\} than the \{reference\}. \\

farther from $X$
& The \{subject\} is farther from the
  \{target\} than the \{reference\}. \\

\bottomrule
\end{tabularx}

\caption{Canonical natural-language reasoning
templates for the supported spatial relations.
Each template verbalizes a spatial predicate
using object references and intermediate results
obtained from spatial query execution. These
templates are used to construct execution-grounded
reasoning traces before linguistic paraphrasing.}
\label{tab:trace-templates-relations}
\end{table*}

\subsection{Filtering Criteria}\label{app:method_filtering}
We apply deterministic filtering to ensure that generated questions have valid intermediate results and that spatial relations in 3D scenes are distinguishable from the rendered image. 

\paragraph{Query validity.} As described in Section~\ref{sec:filtering-trivial-compositions}, we exclude questions with empty intermediate results or redundant spatial compositions. For branching queries, intermediate sets must be distinct. Intersection results must be non-empty and differ from each contributing set, while set-difference results must differ from the first intermediate set.

\paragraph{3D visibility and relation filtering.} We project each 3D object into the camera view and re-rasterize its mesh using a depth buffer to determine its visible pixels and unoccluded silhouette. Table~\ref{tab:3d-filtering} summarizes the filtering criteria.

\begin{table}[t]
\centering
\footnotesize
\setlength{\tabcolsep}{3pt}
\renewcommand{\arraystretch}{1.10}

\begin{tabularx}{\columnwidth}{
@{}
>{\raggedright\arraybackslash}p{0.17\columnwidth}
>{\raggedright\arraybackslash}X
>{\raggedright\arraybackslash}p{0.25\columnwidth}
@{}
}
\toprule
\textbf{Level} & \textbf{Criterion} & \textbf{Threshold} \\
\midrule
Object & Visible image area
    & $\geq 0.25\%$ \\
Object & Unoccluded projected silhouette
    & $\geq 50\%$ \\
\midrule
Relation & Left / right separation
    & $> 0.005W$ \\
Relation & Above / below separation
    & $> 0.005H$ \\
Relation & Front / behind depth separation
    & $> \delta_{AB}$ \\
\bottomrule
\end{tabularx}

\caption{Filtering criteria for 3D scene generation.
$W$ and $H$ denote image width and height.
Directional margins are measured between
coordinate extrema derived from projected
visible boundary polygons. The depth margin
is $\delta_{AB}=0.02\max(z_A,z_B)$, where
$z_A$ and $z_B$ denote the compared camera-frame
depths, increasing with distance from the camera.
All eight object corners must have positive depth.
Object-level filtering precedes relation-level
filtering.}
\label{tab:3d-filtering}
\end{table}

Objects failing any object-level criterion are excluded from scene inventories, reference objects, candidate sets, and answer options. We apply relation-level criteria to eligible object pairs. Only objects and relations satisfying these criteria are used to construct questions.

\subsection{Details on Paraphrasing} \label{app:method_paraphrasing}

\paragraph{Paraphrasing Model Details}
We use Qwen3-32B~\citep{yang-2025-qwen3} to paraphrase canonical questions and execution-grounded reasoning traces. Paraphrasing introduces linguistic variation without modifying the underlying spatial queries, executed intermediate results, or final answers. Table~\ref{tab:paraphrasing-parameters} summarizes the generation settings for both stages.

\begin{table}[t]
\centering
\small
\setlength{\tabcolsep}{4pt}
\renewcommand{\arraystretch}{1.05}
\begin{tabular*}{\columnwidth}{
    @{\extracolsep{\fill}}lcc@{}}
\toprule
\textbf{Hyperparameter}
    & \textbf{Question}
    & \textbf{Reasoning Trace} \\
\midrule
Model & Qwen3-32B & Qwen3-32B \\
Temperature & 0.8 & 0.5 \\
Top-p & 0.95 & 0.95 \\
Max sequence length & 1,024 & 4,096 \\
Variants & 3 & 3 \\
\bottomrule
\end{tabular*}
\caption{Generation settings for question and
reasoning-trace paraphrasing.}
\label{tab:paraphrasing-parameters}
\end{table}

\paragraph{Question Paraphrasing.} \label{app:paraphrasing_prompt}
Given a canonical question, we prompt Qwen3-32B to produce three linguistically diverse variants while preserving its spatial relations, object references, and answer semantics (Figure~\ref{fig:question-paraphrase-prompt}). Variations include changes in vocabulary and sentence structure. The prompt requests a JSON array containing three complete questions.

\begin{figure}[htbp]
    \centering 
    \fcolorbox{black}{gray!5}{\parbox{0.92\linewidth}{    
    \ttfamily \footnotesize
    
    Your task is to rephrase the spatial questions. You are given a spatial reasoning question about objects present in the scene. Your job is to produce 3 diverse rephrasings of the question that preserve EXACTLY the SAME meaning and expected answer, but vary the phrasing, vocabulary, and sentence structure. \\
    
    Rules:

    - Preserve ALL spatial relationships mentioned. DO NOT add, remove, or change them.

    - Vary relationship wording 

    - Vary question structure

    - Produce EXACTLY 3 rephrasings. Output JSON array of strings. Nothing else.

    - Each rephrasing must be a valid, complete question ending with '?'.
}}  
    \caption{System prompt used for paraphrasing the questions.}
    \label{fig:question-paraphrase-prompt}
\end{figure}

\paragraph{Reasoning Step Paraphrasing Prompt}
We separately paraphrase the canonical reasoning traces constructed from executed spatial relation blocks. The prompt (Figure~\ref{fig:reasoning-paraphrase-prompt} ) instructs Qwen3-32B to preserve the logical sequence, spatial predicates, object references, and verified intermediate results while varying their natural-language expressions. Each step remains grounded in its original execution-derived object set.

\begin{figure}[htbp]
    \centering 
    \fcolorbox{black}{gray!5}{\parbox{0.92\linewidth}{    
    \ttfamily \footnotesize
    
    Your task is to rephrase the reasoning-step. Given a list of reasoning steps for a spatial question, rephrase them using varied vocabulary while preserving the logical flow and all spatial information. 

    Rules: 
    
    - DO NOT add, remove, or alter spatial operations, intermediate results, or conclusions.
    
    - The order and dependencies of reasoning steps.
    
    - Every spatial relation and its direction.
    
    - All object identifiers and references.
    
    - All intermediate object sets and their members.
    
    - The final answer and its semantics.
    
    - Produce EXACTLY 3 rephrasings. Output JSON array of strings. Nothing else.    
    }}  
    \caption{System prompt used for paraphrasing reasoning steps}
    \label{fig:reasoning-paraphrase-prompt}
\end{figure}

\paragraph{Role of paraphrasing.}
We apply paraphrasing after spatial query execution and canonical trace construction. The language model does not determine spatial relations or compute intermediate states. The executable scene representation remains the source of the supervision targets, while paraphrasing changes only their linguistic form.

\subsection{Details on Training}
\subsubsection{Training Hyperparameters}
\label{app:training-hyperparameters}
Table~\ref{tab:hparams-sidebyside} summarizes the
hyperparameters for the two-stage training pipeline.
GRPO is initialized from the corresponding
trace-supervised SFT checkpoint.

\begin{table*}[t]
\centering
\small
\renewcommand{\arraystretch}{1.05}

\begin{minipage}[t]{0.48\textwidth}
\vspace{0pt}
\centering
\subcaption{Stage 1: SFT (LLaMA-Factory)}
\label{tab:sft-hparams}

\begin{tabular*}{\linewidth}{@{}p{0.60\linewidth}@{\extracolsep{\fill}}p{0.38\linewidth}@{}}
\toprule
\textbf{Hyperparameter} & \textbf{Value} \\
\midrule
LoRA rank ($r$) & 16 \\
LoRA alpha ($\alpha$) & 32 \\
LoRA dropout & 0.10 \\
Per-device batch size & 2 \\
Gradient accumulation & 4 \\
Precision & bf16 \\
Learning rate & $1\times10^{-4}$ \\
LR scheduler & Cosine \\
Warmup ratio & 0.10 \\
Training questions & 23,800 \\
Epochs & 1 \\
Max sequence length & 8,192 \\
Max image pixels & 1,048,576 \\
\bottomrule
\end{tabular*}
\end{minipage}
\hfill
\begin{minipage}[t]{0.48\textwidth}
\vspace{0pt}
\centering
\subcaption{Stage 2: GRPO (TRL)}
\label{tab:grpo-hparams}

\begin{tabular*}{\linewidth}{@{}p{0.60\linewidth}@{\extracolsep{\fill}}p{0.38\linewidth}@{}}
\toprule
\textbf{Hyperparameter} & \textbf{Value} \\
\midrule
LoRA rank ($r$) & 16 \\
LoRA alpha ($\alpha$) & 32 \\
LoRA dropout & 0.05 \\
Generations per prompt & 6 \\
Per-device batch size & 6 completions \\
Gradient accumulation & 4 \\
Precision & bf16 \\
Learning rate & $2\times10^{-5}$ \\
LR scheduler & Cosine \\
Warmup ratio & 0.05 \\
Training questions & 1,736 \\
Epochs & 2 \\
Max completion length & 384 \\
Max image pixels & 1,048,576 \\
KL coefficient ($\beta$) & 0.02 \\
Answer reward weight & 1.0 \\
Format reward weight & 0.2 \\
\bottomrule
\end{tabular*}
\end{minipage}

\caption{Training hyperparameters for the two-stage
pipeline, with Qwen2.5-VL-3B as the reference
configuration. GRPO is initialized from the
trace-supervised SFT checkpoint. Both stages use
AdamW with LoRA applied to language-model linear
layers while freezing the vision encoder. Each
training run uses one GPU. Qwen2.5-VL-7B and
InternVL3-2B use an SFT per-device batch size of 1
with 8 gradient-accumulation steps, preserving the
effective batch size of 8. Qwen2.5-VL-7B GRPO uses
602,112 maximum image pixels. GRPO samples six
completions per prompt in each of two epochs.
Other settings follow library defaults.}
\label{tab:hparams-sidebyside}
\end{table*}

\subsubsection{Models and Computational Resources} \label{app:training-modelsresources}
We conduct experiments using Qwen2.5-VL-3B, Qwen2.5-VL-7B, Qwen3-VL-4B, and InternVL3-2B. Experiments are performed on a system equipped
with four NVIDIA RTX 6000 GPUs, each with
48 GB of memory. Each training run uses a
single GPU, with separate experiments executed
concurrently across the available GPUs.

For Qwen2.5-VL-3B, the complete training
pipeline, including supervised fine-tuning (SFT)
and group relative policy optimization (GRPO),
requires approximately 15--16 GPU-hours on
one GPU.

%% file: sections/appendix/add_examples.tex
\section{Synthetic Scene Example} \label{app:synthetic-data-example}
We provide representative examples of the 2D map-like and 3D rendered scenes generated by \modelname. Both environments maintain the object geometry required for spatial query execution but differ in the spatial relations they represent and the visual information needed to determine those relations.

\subsection{2D Map-like Scenes} \label{app:synthetic-data-example-2d}
\begin{figure}[h]
    \centering
    \includegraphics[width=\linewidth]{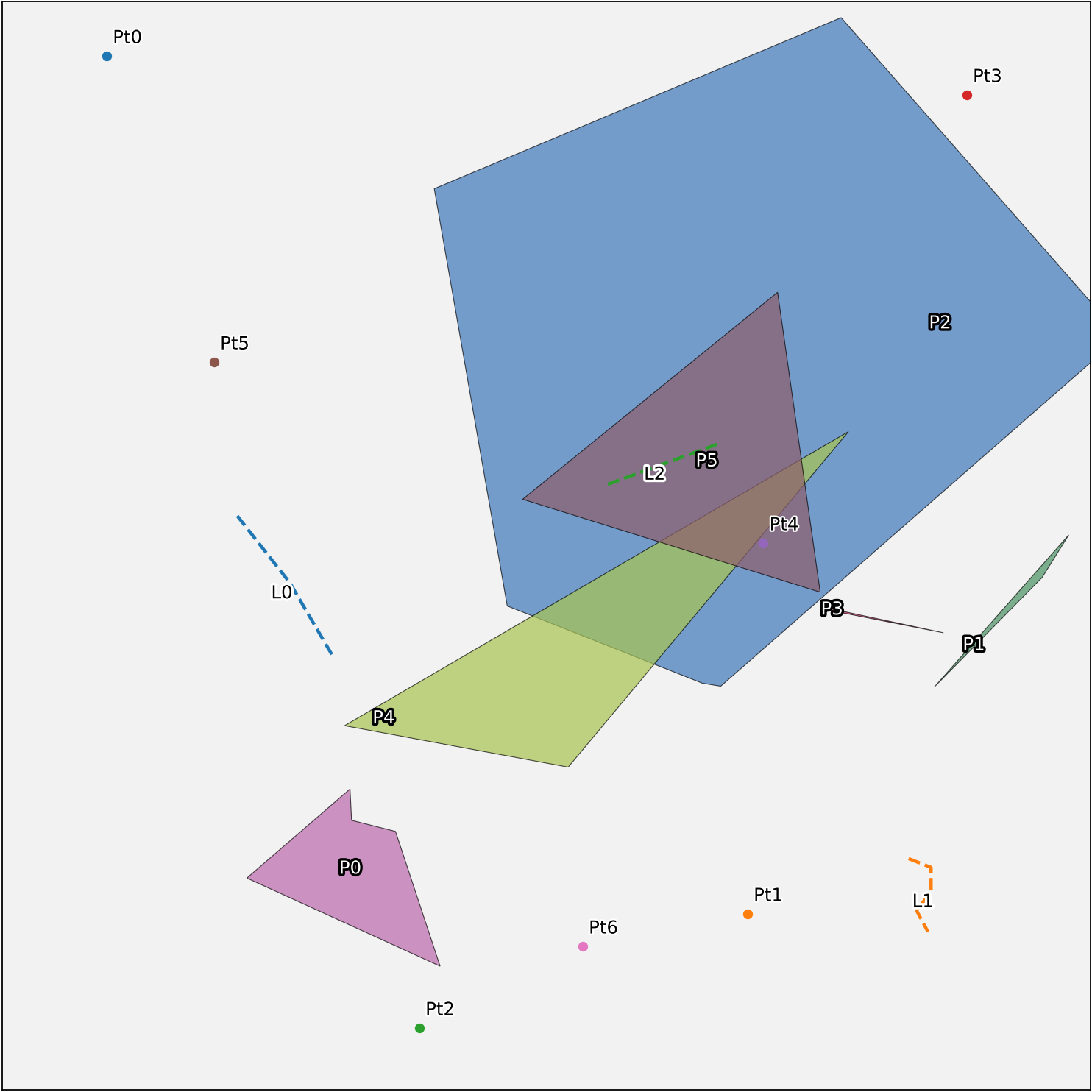}
    \caption{Example 2D map-like scene created by \modelname containing labeled points, lines, and polygons with varying geometric configurations.}
    \label{fig:topological-scene-sample}
\end{figure}

Figure~\ref{fig:topological-scene-sample} shows a generated 2D map-like scene containing labeled points, lines, and polygons with varying positions, shapes, sizes, and colors. The scene includes intersecting and overlapping geometry, allowing questions to involve topological relations such as containment, intersection, and adjacency, as well as directional relations between objects. Since the exact geometries are recorded, \modelname can evaluate these relations and compose multi-step questions without relying on approximate bounding boxes.

\subsection{3D Rendered Scenes} \label{app:synthetic-data-example-3d}
\begin{figure}[h]
    \centering
    \includegraphics[width=\linewidth]{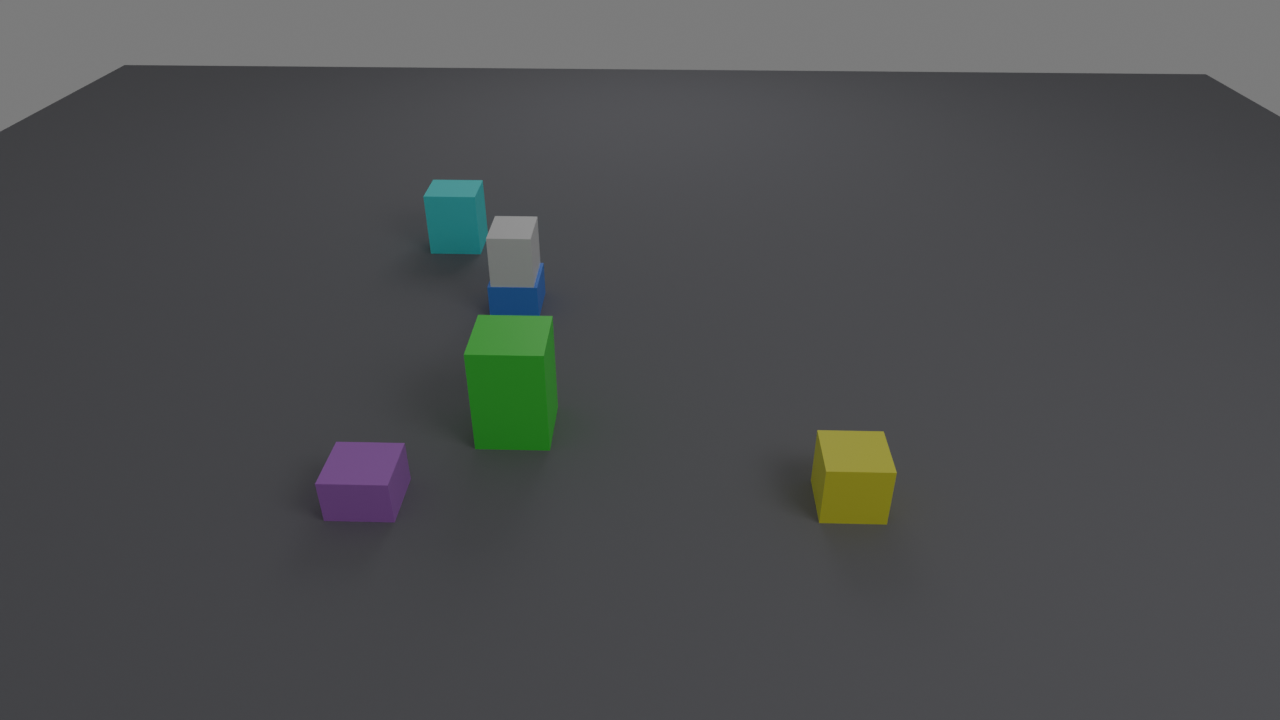}
    \caption{Example 3D rendered scene created by \modelname containing colored geometric objects with viewpoint-dependent spatial arrangements and partial occlusion.}
    \label{fig:spatial-scene-sample}
\end{figure}

Figure~\ref{fig:spatial-scene-sample} shows a generated 3D scene containing colored geometric objects viewed from a specified camera position. Unlike the 2D map-like scenes, the apparent spatial arrangement depends on the perspective projection and object depth. \modelname records each object's 3D geometry, projected visible boundary, camera-frame position, and depth information. These representations support viewpoint-dependent relations such as \textit{left of}, \textit{right of}, \textit{in front of}, \textit {behind}, as well as visibility-based relations involving occlusion. The scene illustrates spatial reasoning that requires both projected image positions and depth information.

\section{Bounding Box vs. Boundary Representation} \label{app:bbox-vs-boundary}

\begin{figure*}[h]
    \centering

    \begin{subfigure}{0.48\textwidth}
        \centering
        \includegraphics[width=\linewidth]{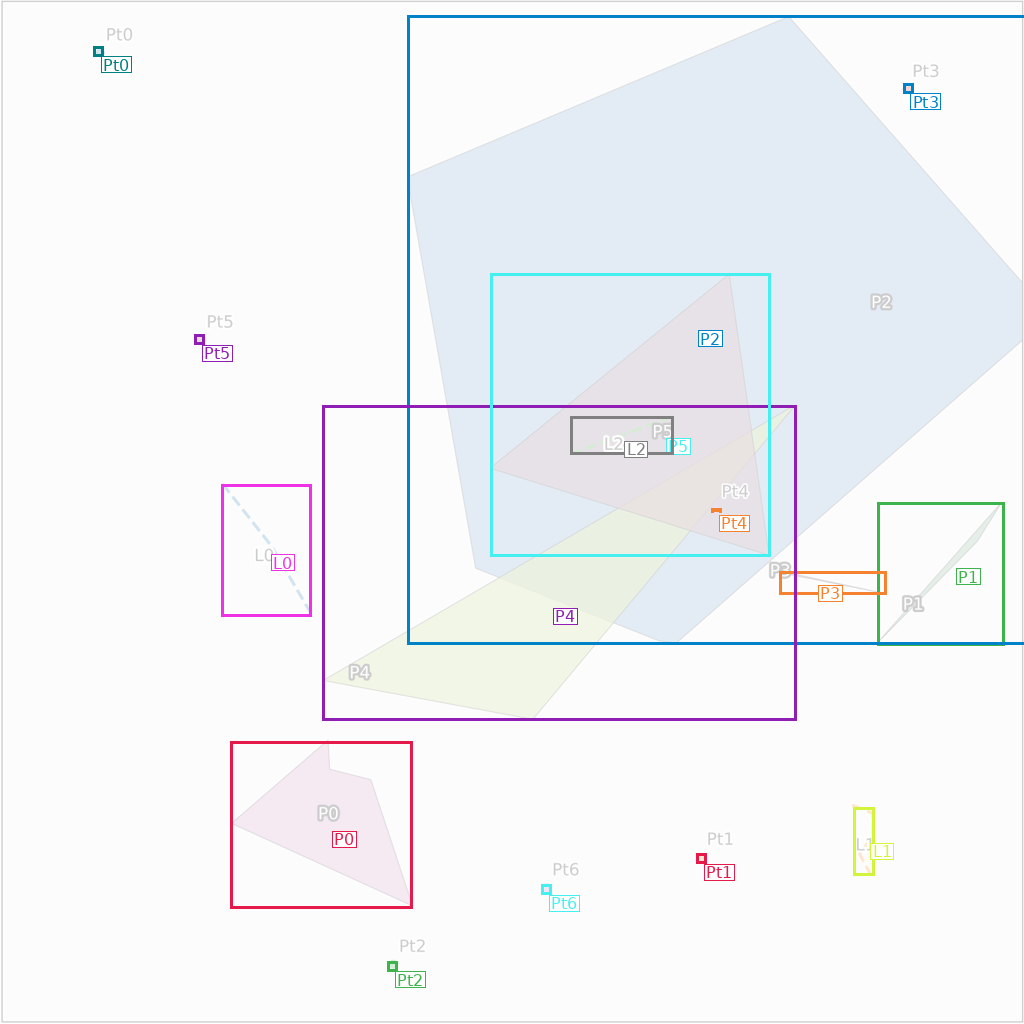}
        \caption{Objects represented by bounding boxes}
        \label{fig:topological-bbox}
    \end{subfigure}
    \hfill
    \begin{subfigure}{0.48\textwidth}
        \centering
        \includegraphics[width=\linewidth]{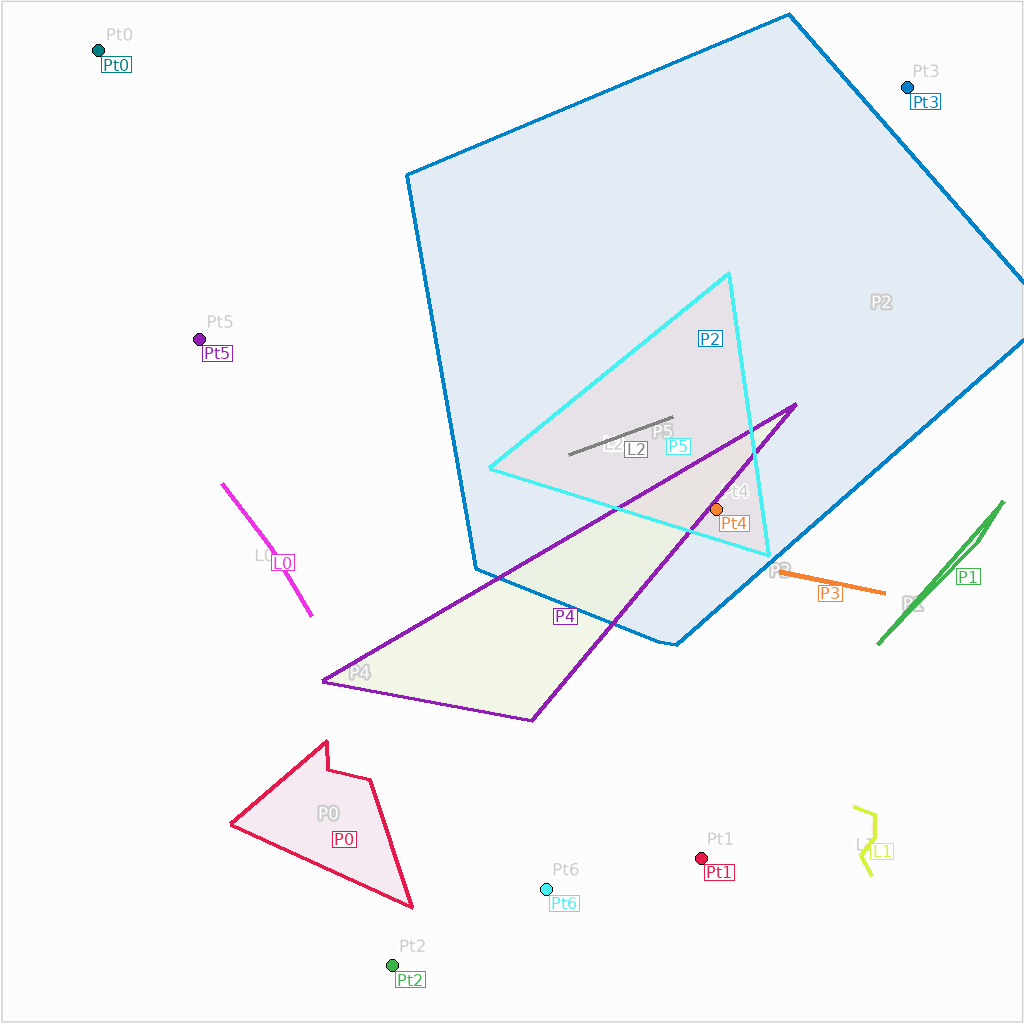}
        \caption{Objects represented by boundary polygons}
        \label{fig:topological-boundary}
    \end{subfigure}
    \caption{Comparison of bounding boxes and exact object boundaries in a 2D scene. Bounding boxes can introduce false spatial relationships by covering regions outside the actual object geometries.}
    \label{fig:topological-bbox-boundary}
\end{figure*}

Bounding boxes provide approximate object localization but do not preserve the geometric boundaries needed to precisely evaluate spatial relations. In 2D scenes, axis-aligned bounding boxes may include regions outside an object's actual geometry, introducing false containment or intersection relations. Figure~\ref{fig:topological-bbox-boundary} illustrates this issue. Although polygons P1 (green polygon) and P3 (orange polygon) are geometrically disjoint, their bounding boxes overlap, which could incorrectly suggest that the polygons intersect. Using exact object boundaries instead allows \modelname to distinguish these configurations and evaluate topological relations, including containment, intersection, adjacency, and disjointness.

\begin{figure*}[h]
    \centering
    \begin{subfigure}{0.48\textwidth}
        \centering
        \includegraphics[width=\linewidth]{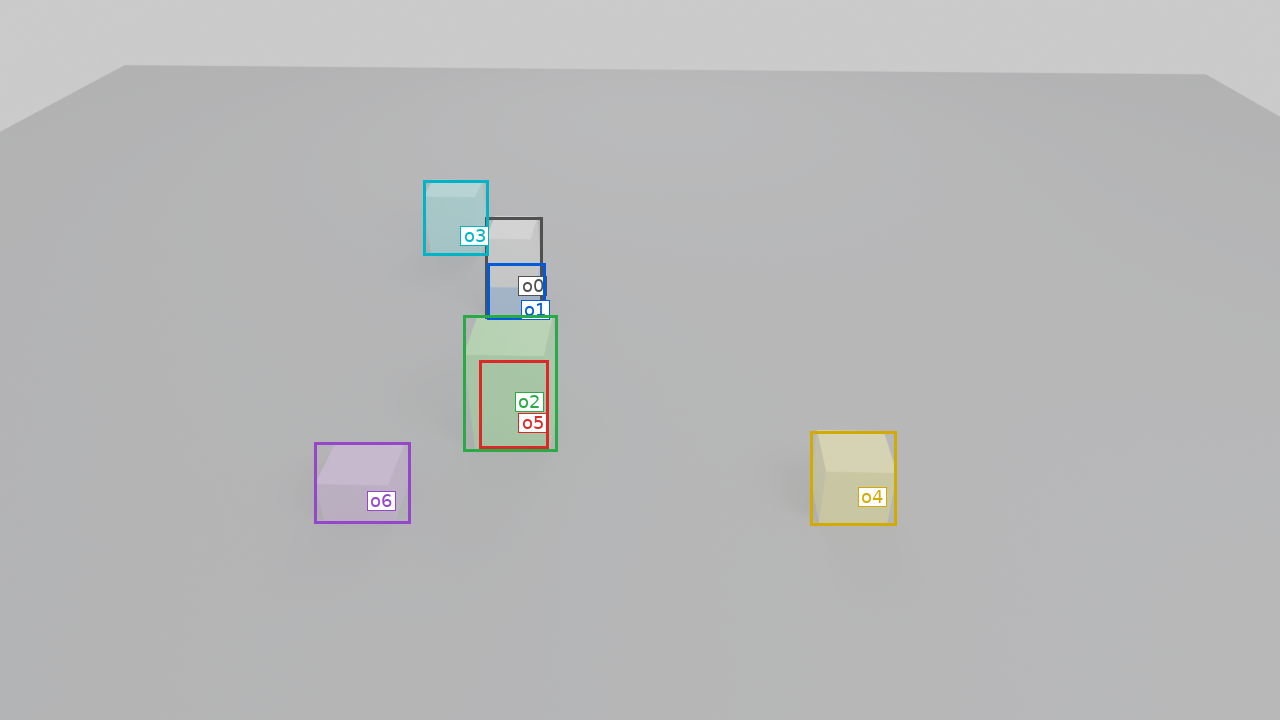}
        \caption{Objects represented by bounding boxes}
        \label{fig:spatial-bbox}
    \end{subfigure}
    \hfill
    \begin{subfigure}{0.48\textwidth}
        \centering
        \includegraphics[width=\linewidth]{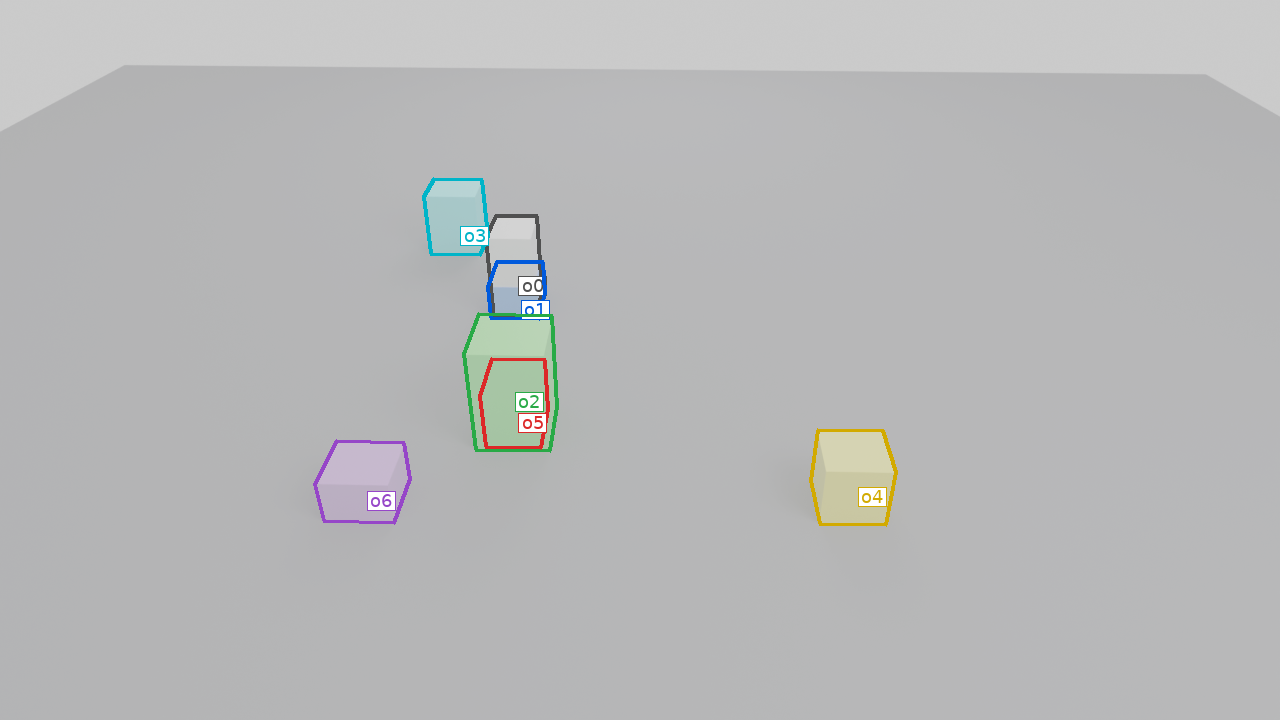}
        \caption{Objects represented by boundary polygons}
        \label{fig:spatial-boundary}
    \end{subfigure}
    \caption{Comparison of bounding boxes and projected visible boundaries in a 3D rendered scene. Projected boundaries preserve object silhouettes for more precise evaluation of image-plane spatial relations.}
    \label{fig:spatial-bbox-boundary}
\end{figure*}

A similar issue arises in 3D rendered scenes, where spatial relations depend on projected shapes and camera-relative object positions. As shown in Figure~\ref{fig:spatial-bbox-boundary}, bounding boxes discard information about visible object silhouettes and can introduce apparent overlap between objects whose projected boundaries do not overlap. For example, the projected boundary of object o5 lies within that of o2, but their bounding boxes cannot distinguish this configuration from partial overlap between the two projected shapes. Moreover, projected boundaries alone do not establish depth ordering. Therefore, \modelname uses projected visible boundaries to evaluate image-plane relations and camera-frame depth information for depth-dependent relations, rather than relying solely on bounding-box approximations.

\section{End-to-End Trace Creation Sample} \label{app:e2e-sample}

\lstdefinestyle{sqlstyle}{
    basicstyle=\ttfamily\scriptsize,
    breaklines=true,
    columns=fullflexible,
    keepspaces=true,
    frame=single,
    rulecolor=\color{black!25},
    backgroundcolor=\color{black!3},
    showstringspaces=false
}

\lstdefinestyle{tracestyle}{
    basicstyle=\ttfamily\scriptsize,
    breaklines=true,
    columns=fullflexible,
    keepspaces=true,
    frame=single,
    rulecolor=\color{black!25},
    backgroundcolor=\color{black!3},
    showstringspaces=false
}

We illustrate how \modelname constructs an execution-grounded reasoning trace from a synthetic scene and a multi-step spatial question. The example consists of two spatial relation blocks, each evaluated against the recorded object geometries. Their executed results provide the intermediate states used to construct the natural-language reasoning trace and final answer.

\subsection{Synthetic Scene and Question}

\begin{figure}[h!]
    \centering
    \includegraphics[width=\linewidth]{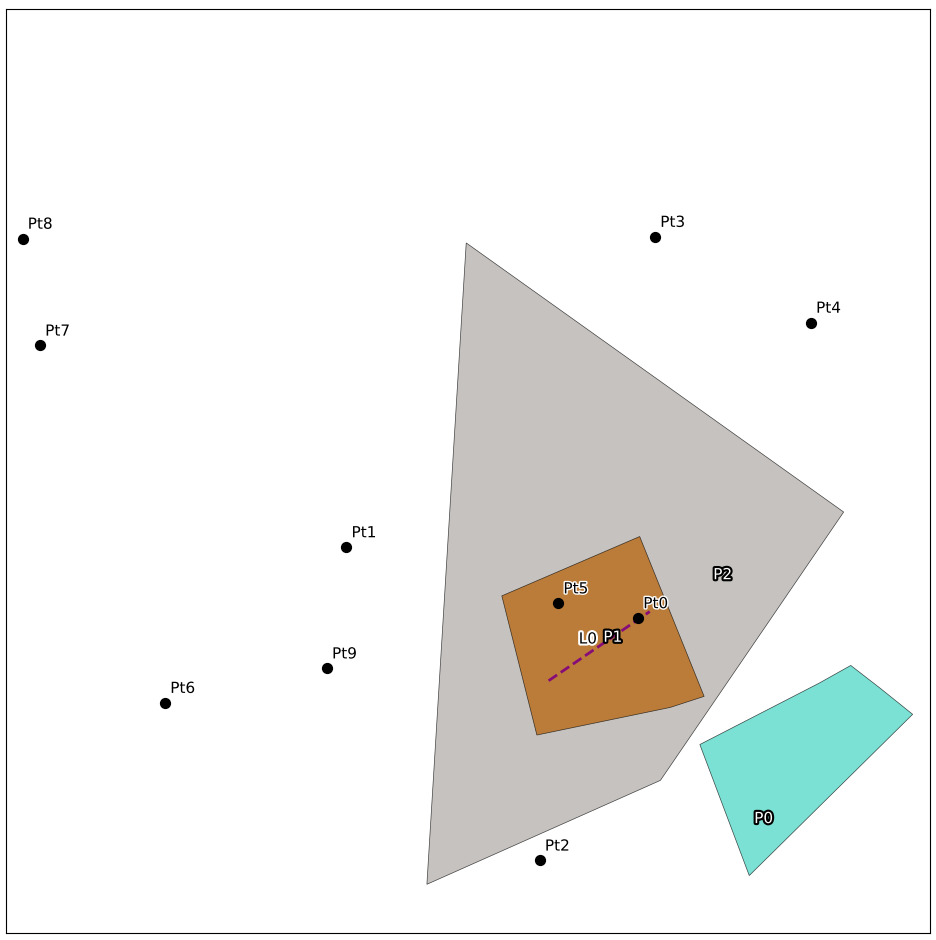}
    \caption{Sample synthetic topological scene}
    \label{fig:e2e-synthetic-image}
\end{figure}

Figure~\ref{fig:e2e-synthetic-image} shows a generated 2D map-like scene containing labeled points, lines, and polygons. We consider the following question:
\begin{quote}
\textit{Which point is within polygon within P2?}\footnote{As questions are initially created from templates, they may contain grammatical artifacts, including imperfect singular-plural agreement.}
\end{quote}

Answering this question requires two spatial operations. First, identify the polygons contained within P2. Then, identify the points contained within the resulting polygons.

\subsection{Spatial Query Execution} \label{example:execution}

\modelname represents each object using its recorded geometry and attributes in the \texttt{generated\_geometries} table. We use PostGIS to execute spatial queries over these geometries. Listing~\ref{lst:e2e-spatial-sql} presents the query corresponding to the example.

\begin{lstlisting}[style=sqlstyle, language=SQL, caption={Spatial SQL query for the end-to-end topological reasoning example.}, label={lst:e2e-spatial-sql}]
SELECT DISTINCT
    t_final.id,
    t_final.name
FROM generated_geometries AS t_final
JOIN generated_geometries AS t1
    ON ST_Within(t_final.geom, t1.geom)
JOIN generated_geometries AS t2
    ON ST_Within(t1.geom, t2.geom)
WHERE t_final.type = 'point'
  AND t1.type = 'polygon'
  AND t2.type = 'polygon'
  AND t2.name = 'P2'
ORDER BY t_final.id;
\end{lstlisting}

The query composes two containment predicates. The first \texttt{ST\_Within(t1.geom, t2.geom)} selects polygons contained within P2. The second \texttt{ST\_Within(t\_final.geom, t1.geom)} selects points contained within the previously identified polygon. The final query returns the distinct points satisfying both relations.

\subsection{Execution Plan and Verified Intermediate States} \label{app:example-executionplan}
\begin{figure}[h!]
    \centering
    \includegraphics[width=\linewidth]{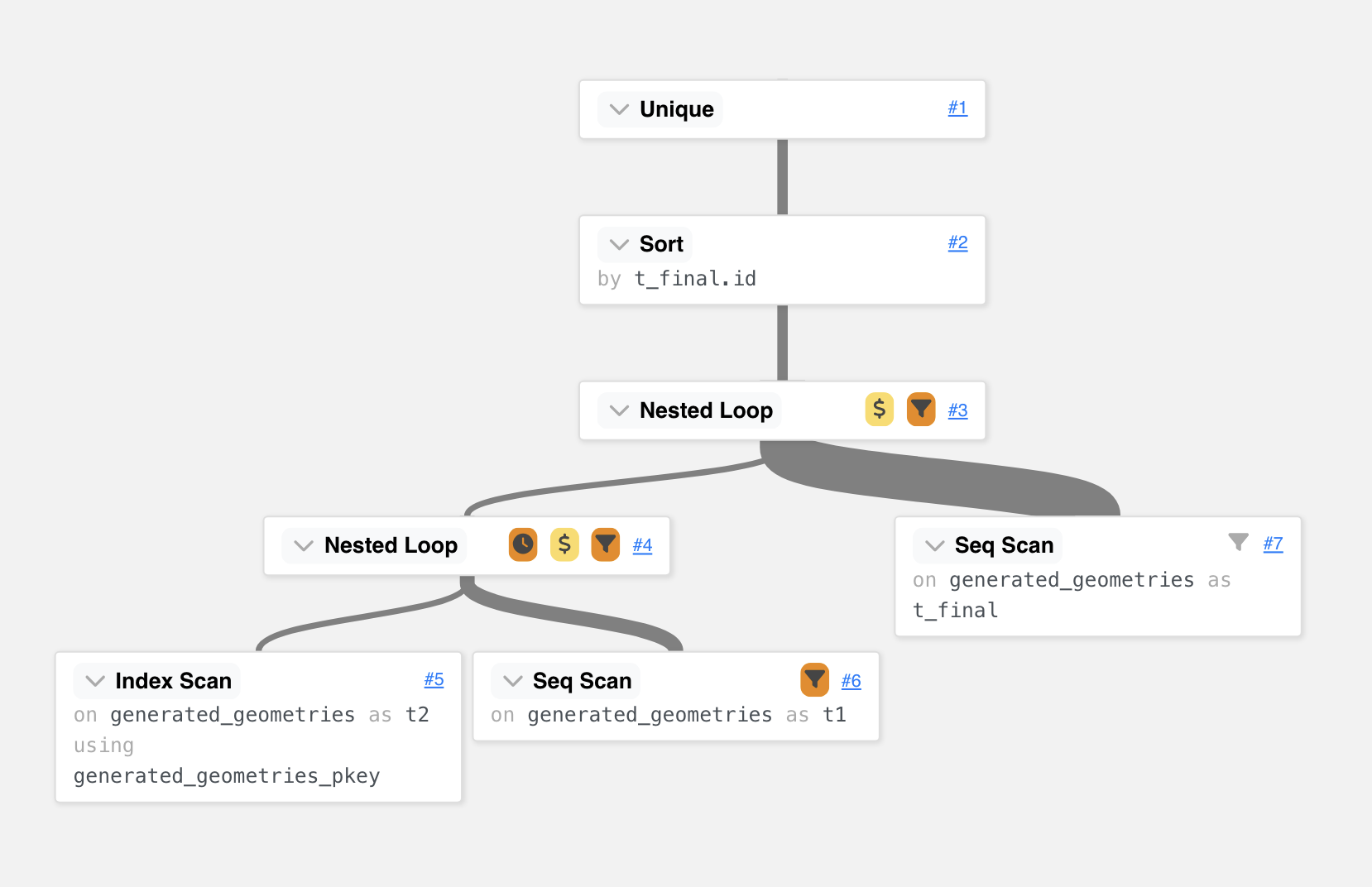}
    \caption{Visualized execution plan for the spatial query. The plan shows how the database retrieves P2, scans candidate polygons and points, applies spatial containment predicates, sorts the resulting points, and removes duplicates.}
    \label{fig:e2e-trace-visualization}
\end{figure}

After executing the spatial SQL query, we retrieve its optimized execution plan from PostGIS. The plan provides a finer-grained breakdown of the operations used to evaluate the question, including object retrieval, spatial joins, sorting, and duplicate removal. Listing~\ref{lst:e2e-execution-plan} presents the execution plan, and Figure~\ref{fig:e2e-trace-visualization}
visualizes its structure.

\begin{lstlisting}[style=tracestyle, caption={Execution plan for the spatial SQL query.}, label={lst:e2e-execution-plan}]
Unique
  -> Sort
       Sort Key: t_final.id

       -> Nested Loop
            Join Filter: ST_Within(t_final.geom, t1.geom)

            -> Nested Loop
                 Join Filter: ST_Within(t1.geom, t2.geom)

                 -> Index Scan on generated_geometries as t2
                      Index Cond: name = 'P2'
                      Filter: type = 'polygon'

                 -> Seq Scan on generated_geometries as t1
                      Filter: type = 'polygon'

            -> Seq Scan on generated_geometries as t_final
                 Filter: type = 'point'
\end{lstlisting}

We use this plan to identify the operations associated with each spatial relation and obtain their intermediate results through the corresponding spatial subqueries. For the example, the execution first retrieves the reference polygon P2 and identifies candidate polygons. Evaluating the containment predicate returns the intermediate object set \textit{P1}. The subsequent spatial operation retrieves candidate points and evaluates their containment within P1, producing \textit{Pt5} and \textit{Pt0}, which are sorted to give the final answer set \textit{{Pt0, Pt5}}.

Listing~\ref{lst:e2e-operation-reasoning} presents the resulting operation-level trace. We obtain the intermediate object sets through spatial query execution rather than generating them with an LVLM. This provides an execution-grounded record from which we can construct natural-language reasoning steps.

\begin{lstlisting}[style=tracestyle, caption={Operation-level reasoning trace derived from the execution plan.}, label={lst:e2e-operation-reasoning}]
1. Retrieve reference polygon P2.
2. Retrieve candidate polygons.
3. Identify polygons within P2.
   Result: {P1}.
4. Retrieve candidate points.
5. Identify points within P1.
   Result: {Pt5, Pt0}.
6. Sort the resulting points by ID.
   Result: {Pt0, Pt5}.
7. Remove duplicate points.
   Final result: {Pt0, Pt5}.
\end{lstlisting}

\subsection{Execution to Natural-Language Reasoning} \label{app:example-verbalization}

The operation-level trace provides a detailed record of how the spatial query derives its answer. However, not all database operations correspond to spatial reasoning steps useful for LVLM supervision. For example, sorting and duplicate removal determine how the query returns results but do not introduce additional spatial relations.

Therefore, we convert the operation-level trace into a canonical reasoning trace by retaining the spatial predicates, their reference objects, and the intermediate results they produce, while omitting database-specific operations that do not contribute to spatial reasoning. For the example above, the resulting trace contains two spatial reasoning steps. The first identifies P1 as the polygon contained within P2, and the second identifies Pt0 and Pt5 as the points contained within P1. This preserves the dependency between the two spatial relations while removing unnecessary database implementation details.

\subsection{Paraphrased Reasoning Trace} \label{app:example-paraphrase}

As the final step, \modelname paraphrases the operation-level trace into a natural explanation for LVLM supervision. The paraphrased trace removes database-specific operations while preserving the spatial reasoning structure and verified intermediate results.

\begin{lstlisting}[style=tracestyle, caption={Paraphrased natural-language reasoning trace used for model supervision.}, label={lst:e2e-paraphrased-trace}]
Question: Which points are within the polygon that is within P2?

Reasoning:
First, locate polygon P2 in the image. Then, identify the polygons contained within P2. The polygon P1 is within P2. Next, find the points contained within P1. The points Pt5 and Pt0 are within P1.

<answer>Pt0, Pt5</answer>
\end{lstlisting}

%% file: sections/appendix/add_results.tex
\color{black}
\section{Detailed Results} \label{app:detailed-results}
We report category-level results for the five external spatial reasoning benchmarks, complementing the overall scores in Section~\ref{sec:experiment-results}.

\subsection{2D Spatial Reasoning}\label{app:additional-2d-results}

\paragraph{SpatialEval and FRIEDA.}
Table~\ref{tab:external_2d_results} reports results across the four SpatialEval subsets and three FRIEDA reasoning categories. On SpatialEval, \modelname-Reason+GRPO improves over the corresponding base checkpoints in overall accuracy across all four backbones. Improvements are particularly pronounced on the Maze subset for the Qwen-based models, although performance on the Grid subset does not consistently improve. On FRIEDA, directional reasoning shows the largest gains, while improvements in topology and distance reasoning are less consistent. 

\begin{table*}[h!]
    \centering
    \small
    \begin{tabular}{lccccccccc}
    \toprule
        \multirow{2}{*}{Model Variant}
        & \multicolumn{5}{c}{\textbf{SpatialEval}}
        & \multicolumn{4}{c}{\textbf{FRIEDA}} \\
        \cmidrule(lr){2-6} \cmidrule(lr){7-10}
        & Map & Grid & Maze & Real & Overall
        & Topology & Distance & Direction & Overall \\
    \midrule

        \multicolumn{9}{l}{\textbf{Qwen2.5-VL-3B-based models}} \\
        Base & 65.60 & 71.27 & \underline{33.13} & 57.04 & 56.76 & 11.02 & \textbf{20.00} & 38.55 & 16.40 \\
        SpatialLadder & 50.40 & \underline{69.13} & 27.80 & \underline{61.48} & 52.20 & 11.56 & \textbf{20.00} & 32.53 & 15.80 \\
        SpatialThinker & 64.07 & 71.33 & 32.87 & 57.04 & 56.33 & 12.37 & 17.78 & 36.14 & 16.80 \\
        SenseNova-SI-1.1 & \underline{68.07} & \textbf{72.80} & 31.07 & 60.00 & 57.98 & \underline{13.71} & 17.78 & \underline{42.17} & \underline{18.80} \\
        \modelname-Reason+GRPO & \textbf{68.13} & 68.00 & \textbf{42.60} & \textbf{62.22} & \textbf{60.24} & \textbf{14.78} & 11.11 & \textbf{46.99}  & \textbf{19.80} \\

    \midrule
        \multicolumn{9}{l}{\textbf{Qwen2.5-VL-7B-based models}} \\
        Base & 66.93 & 78.33 & 22.27 & 60.74 & 57.07 & 13.98 & 11.11 & 40.96 & 18.20 \\
        SpatialThinker & 69.27 & 79.93 & 28.87 & \textbf{65.93} & 61.00 & \textbf{14.78} & \textbf{17.78} & 40.96 & 19.40 \\
        SenseNova-SI-1.1 & \textbf{71.40} & \textbf{83.00} & \textbf{62.13} & 64.44 & \textbf{70.24} & \textbf{14.78} & \underline{13.33} & \underline{49.40} & \underline{20.40} \\
        ViLaSR & 66.27 & \underline{80.33} & 34.47 & 64.44 & 61.38 & 12.37 & 0.00 & 33.73 & 14.80 \\
        \modelname-Reason+GRPO & \underline{71.20} & 80.20 & \underline{50.93} & \textbf{65.93} & \underline{67.06} & \textbf{14.78} & \underline{13.33} & \textbf{56.63} & \textbf{21.60} \\

    \midrule
        \multicolumn{9}{l}{\textbf{Qwen3-VL-4B-based models}} \\
        Base & \underline{59.20} & \underline{76.87} & \underline{29.53} & 65.93 & \underline{57.88} & \underline{11.56} & \underline{17.78} & 15.66 & 12.80 \\
        MultiHopSpatial & 57.13 & \textbf{77.93} & 24.87 & \textbf{67.41} & 56.84 & 11.29 & \textbf{22.22} & \underline{19.28} & \underline{13.60} \\
        \modelname-Reason+GRPO & \textbf{68.87} & 71.60 & \textbf{40.73} & \textbf{67.41} & \textbf{62.15} & \textbf{14.52} & 15.56 & \textbf{51.81} & \textbf{20.80} \\

    \midrule
        \multicolumn{9}{l}{\textbf{InternVL3-2B-based models}} \\
        Base & 36.33 & 2.40 & 12.87 & 47.41 & 24.79 & \textbf{11.84} & \underline{10.99} & 39.77 & \textbf{16.60} \\
        SenseNova-SI-1.1 & \underline{39.40} & \textbf{52.40} & \textbf{66.27} & \underline{55.56} & \textbf{53.41} & \underline{9.68} & 8.89 & \textbf{40.96} & \underline{14.80} \\
        \modelname-Reason+GRPO & \textbf{49.27} & \underline{47.47} & \underline{26.47} & \textbf{73.33} & \underline{49.13} & 8.41 & \textbf{12.09} & \underline{40.91} & \underline{14.80} \\
    \bottomrule
    \end{tabular}
    \caption{
    Detailed results on SpatialEval and FRIEDA. SpatialEval scores are reported across Map, Grid, Maze, and Real subsets, while FRIEDA scores are reported across Topology, Distance, and Direction categories. All values indicate accuracy (\%).
    }
    \label{tab:external_2d_results}
\end{table*}

\subsection{3D Spatial Reasoning}\label{app:additional-3d-results}

\paragraph{CV-Bench and BLINK.}
Table~\ref{tab:external_spatial_results} presents the category-level results for CV-Bench and BLINK. On CV-Bench, \modelname-Reason+GRPO improves relation accuracy over the corresponding base checkpoints across all four backbones, whereas improvements in counting, depth, and distance vary by model. The Qwen2.5-VL-3B model achieves notable gains in relation and distance reasoning, while Qwen3-VL-4B shows a slight decrease in overall CV-Bench accuracy despite improved relation accuracy.

\begin{table*}[h!]
    \centering
    \small
    \begin{tabular}{lcccccccc}
    \toprule
        \multirow{3}{*}{Model Variant}
        & \multicolumn{5}{c}{\textbf{CV-Bench}}
        & \multicolumn{3}{c}{\textbf{BLINK}} \\
        \cmidrule(lr){2-6} \cmidrule(lr){7-9}
        & \multicolumn{2}{c}{2D}
        & \multicolumn{2}{c}{3D}
        & \multirow{2}{*}{Overall}
        & \multirow{2}{*}{Spatial}
        & \multirow{2}{*}{Depth}
        & \multirow{2}{*}{Overall} \\
        \cmidrule(lr){2-3} \cmidrule(lr){4-5}
        & Count & Relation & Depth & Distance & & & & \\
    \midrule

        \multicolumn{9}{l}{\textbf{Qwen2.5-VL-3B-based models}} \\
        Base & 66.24 & 77.38 & 74.83 & 69.17 & 71.24 & 81.82 & 74.19 & 78.01 \\
        SpatialLadder & 67.13 & \underline{85.23} & 80.67 & 73.83 & 75.87 & 77.62 & 76.61 & 77.12 \\
        SpatialThinker & \underline{68.40} & 78.31 & 77.67 & 71.67 & 73.42 & 79.72 & 78.23 & 78.97 \\
        SenseNova-SI-1.1 & 67.13 & 81.85 & \textbf{85.00} & \underline{75.50} & \underline{76.65} & \textbf{83.92} & \underline{82.26} & \underline{83.09} \\
        \modelname-Reason+GRPO & \textbf{69.42} & \textbf{87.69} & \underline{83.83} & \textbf{82.50} & \textbf{80.16} & \underline{83.22} & \textbf{85.48} & \textbf{84.35} \\

    \midrule
        \multicolumn{9}{l}{\textbf{Qwen2.5-VL-7B-based models}} \\
        Base & 67.01 & 88.62 & 82.33 & 79.67 & 78.50 & \underline{89.51} & 79.84 & 84.67 \\
        SpatialThinker & \underline{69.29} & \underline{89.69} & 82.33 & 81.67 & 79.91 & \textbf{90.91} & 79.84 & \underline{85.37} \\
        SenseNova-SI-1.1 & 68.02 & 85.38 & \textbf{85.17} & \underline{84.00} & \underline{79.92} & 88.81 & \textbf{87.10} & \textbf{87.95} \\
        ViLaSR & 67.01 & 86.77 & 75.67 & 74.50 & 75.18 & \underline{89.51} & \underline{80.65} & 85.08 \\
        \modelname-Reason+GRPO & \textbf{70.94} & \textbf{92.31} & \underline{83.00} & \textbf{84.50} & \textbf{81.86} & 88.11 & \underline{80.65} & 84.38 \\

    \midrule
        \multicolumn{9}{l}{\textbf{Qwen3-VL-4B-based models}} \\
        Base & 67.01 & \underline{90.00} & \textbf{93.67} & \textbf{87.17} & \textbf{83.59} & \underline{86.71} & \underline{81.45} & \underline{84.08} \\
        MultiHopSpatial & \underline{67.13} & 86.31 & \underline{92.17} & \underline{84.83} & 81.94 & 83.92 & \underline{81.45} & 82.68 \\
        \modelname-Reason+GRPO & \textbf{67.77} & \textbf{93.85} & 91.83 & 84.67 & \underline{83.55} & \textbf{87.41} & \textbf{87.10} & \textbf{87.25} \\

    \midrule
        \multicolumn{9}{l}{\textbf{InternVL3-2B-based models}} \\
        Base & \underline{69.92} & 74.00 & \textbf{85.00} & 70.00 & 74.41 & \textbf{76.92} & \underline{78.23} & \underline{77.57} \\
        SenseNova-SI-1.1 & 66.50 & \textbf{90.15} & \underline{80.33} & \underline{75.83} & \underline{77.39} & 72.73 & 75.00 & 73.86 \\
        \modelname-Reason+GRPO & \textbf{73.98} & \underline{83.54} & 79.67 & \textbf{81.33} & \textbf{79.20} & \underline{76.22} & \textbf{81.45} & \textbf{78.84} \\
    \bottomrule
    \end{tabular}
    \caption{
    Detailed results on CV-Bench and BLINK. CV-Bench includes 2D counting and relation tasks and 3D depth and distance tasks. BLINK includes spatial relation and depth subsets. All values indicate accuracy (\%).
    }
    \label{tab:external_spatial_results}
\end{table*}

\paragraph{MultiHopSpatial}\label{app:additional-multihop-results}
Table~\ref{tab:multihop_spatial_results} reports results by reasoning depth (1-hop, 2-hop, and 3-hop) and viewpoint (egocentric and exocentric). For all four \modelname-Reason+GRPO backbones, accuracy decreases as the number of reasoning hops increases, indicating that longer spatial compositions remain challenging. Exocentric questions also consistently achieve higher accuracy than egocentric questions. Nevertheless, \modelname-Reason+GRPO improves overall MultiHopSpatial accuracy over the corresponding base checkpoints across all four backbones.

We additionally report the MultiHopSpatial-specialized Qwen3-VL-4B checkpoint. Although this model achieves higher accuracy on its corresponding benchmark, it uses benchmark-specific training data and therefore is not included as a comparable external-transfer baseline in the main results.

\begin{table*}[h!]
    \centering
    \small
    \begin{tabular}{lccccccccc}
    \toprule
        \multirow{2}{*}{Model Variant}
        & \multicolumn{3}{c}{\textbf{1-Hop}}
        & \multicolumn{3}{c}{\textbf{2-Hop}}
        & \multicolumn{3}{c}{\textbf{3-Hop}} \\
        \cmidrule(lr){2-4}
        \cmidrule(lr){5-7}
        \cmidrule(lr){8-10}
        & Ego & Exo & Overall
        & Ego & Exo & Overall
        & Ego & Exo & Overall \\
    \midrule

        \multicolumn{10}{l}{\textbf{Qwen2.5-VL-3B-based models}} \\
        Base
        & 22.80 & 54.40 & 38.60 & 19.20 & 50.67 & 34.93 & 16.53 & \underline{39.60} & \underline{28.07} \\
        
        SpatialLadder
        & 22.27 & 36.00 & 29.13 & \underline{21.87} & 38.80 & 30.33 & \textbf{22.13} & 28.00 & 25.07 \\
        
        SpatialThinker
        & 24.13 & \underline{56.53} & 40.33 & 20.53 & \underline{51.73} & \underline{36.13} & 15.60 & 38.93 & 27.27 \\
        
        SenseNova-SI-1.1
        & \underline{27.87} & 55.60 & \underline{41.73} & 17.33 & 46.80 & 32.07 & \underline{17.60} & 36.67 & 27.13 \\
        
        \modelname-Reason+GRPO
        & \textbf{32.80} & \textbf{66.00} & \textbf{49.40} & \textbf{23.60} & \textbf{55.20} & \textbf{39.40} & 16.13 & \textbf{45.47} & \textbf{30.80} \\

    \midrule
        \multicolumn{10}{l}{\textbf{Qwen2.5-VL-7B-based models}} \\
        Base
        & 20.40 & 61.60 & 41.00 & 17.33 & 54.53 & \underline{35.93} & 13.60 & \underline{38.00} & \underline{25.80} \\
        
        SpatialThinker
        & \underline{21.60} & \underline{62.93} & \underline{42.27} & 16.40 & \underline{54.93} & 35.67 & 13.33 & 37.07 & 25.20 \\
        
        SenseNova-SI-1.1
        & 21.47 & 61.47 & 41.47 & \underline{19.47} & 49.07 & 34.27 & 12.27 & 30.27 & 21.27 \\
        
        ViLaSR
        & 16.67 & 49.87 & 33.27 & 14.93 & 43.87 & 29.40 & \textbf{15.20} & 30.27 & 22.73 \\
        
        \modelname-Reason+GRPO
        & \textbf{25.47} & \textbf{68.00} & \textbf{46.73} & \textbf{21.47} & \textbf{57.87} & \textbf{39.67} & \underline{14.93} & \textbf{42.13} & \textbf{28.53} \\

    \midrule
        \multicolumn{10}{l}{\textbf{Qwen3-VL-4B-based models}} \\
        Base
        & 26.00 & 69.33 & 47.67 & 21.47 & 50.80 & 36.13 & \underline{15.73} & 38.93 & 27.33 \\
        
        MultiHopSpatial
        & \textbf{64.27} & \textbf{82.40} & \textbf{73.33} & \textbf{54.00} & \textbf{72.67} & \textbf{63.33} & \textbf{44.67} & \textbf{56.40} & \textbf{50.53} \\
        
        \modelname-Reason+GRPO
        & \underline{30.40} & \underline{74.27} & \underline{52.33} & \underline{22.27} & \underline{53.47} & \underline{37.87} & 15.20 & \underline{45.07} & \underline{30.13} \\

    \midrule
        \multicolumn{10}{l}{\textbf{InternVL3-2B-based models}} \\
        Base
        & 26.00 & 53.73 & 39.87 & \underline{22.40} & 51.20 & 36.80 & {19.33} & 36.93 & 28.13 \\
        
        SenseNova-SI-1.1
        & \underline{27.87} & \underline{63.20} & \underline{45.53} & \textbf{25.20} & \textbf{56.67} & \textbf{40.93} & 18.93 & \textbf{47.07} & \textbf{33.00} \\
        
        \modelname-Reason+GRPO
        & \textbf{30.67} & \textbf{67.07} & \textbf{48.87} & 20.13 & \underline{54.93} & \underline{37.53} & \textbf{21.20} & \underline{44.67} & \underline{32.93} \\

    \bottomrule
    \end{tabular}
    \caption{
    Detailed results on MultiHopSpatial across reasoning depths and viewpoints. Ego and Exo denote egocentric and exocentric questions, respectively. The MultiHopSpatial-specialized checkpoint uses benchmark-specific training and is included for reference only. All values indicate accuracy (\%).
    }
    \label{tab:multihop_spatial_results}
\end{table*}

\section{Statistical Significance Testing} \label{app:statistical-significance}

We assess the statistical significance of external benchmark results using 10,000 paired bootstrap resamples. For each backbone and benchmark, we compare \modelname-Reason+GRPO against the strongest competing model in the corresponding backbone group. Each bootstrap iteration samples question indices with replacement, using identical sampled indices for both models to preserve the paired evaluation setting.

For composite benchmarks (CV-Bench, BLINK, and SpatialEval), we resample question indices with replacement independently within each evaluation subset, preserving the original subset sizes. We use identical sampled indices for both models to maintain paired comparisons. For each bootstrap iteration, we recompute the overall score from the resampled subset accuracies using the benchmark's original aggregation weights. Other benchmarks follow their respective scoring procedures.

We calculate the difference in overall accuracy
between \modelname-Reason+GRPO and the competing
model for each bootstrap resample. We estimate the two-sided $p$-value, and consider differences statistically significant when $p<0.05$. The main results table highlights significant improvements and distinguishes lower scores that are not statistically significant.

%% file: sections/appendix/add_licenses.tex
\color{black}
\section{Licenses and Intended Use} \label{app:licenses}

\paragraph{Generated data.}
\modelname generates synthetic 2D and 3D scenes, spatial questions, execution-derived answers, and natural-language reasoning traces. The synthetic scenes are constructed from geometric primitives rather than existing benchmark images. These artifacts support research on multimodal spatial reasoning and execution-grounded supervision. Any public release of the generated data and code will specify its distribution license separately from the licenses of third-party resources.

\paragraph{Pretrained models.}
We use publicly available pretrained models for fine-tuning, evaluation, and linguistic diversification. Qwen2.5-VL-7B-Instruct~\citep{bai-2025-qwen25vl}, Qwen3-VL-4B-Instruct~\citep{bai-2025-qwen3vl}, InternVL3-2B~\citep{zhu-2025-internvl3}, and Qwen3-32B~\citep{yang-2025-qwen3} are distributed under the Apache-2.0 license. Qwen2.5-VL-3B-Instruct~\citep{bai-2025-qwen25vl} is distributed under the Qwen Research License Agreement. All pretrained models and fine-tuned derivatives remain subject to their applicable upstream licensing terms.

\paragraph{Specialized baseline models.}
We evaluate publicly released checkpoints of SpatialLadder~\citep{li-2025-spatialladder}, SpatialThinker~\citep{batra-2025-spatialthinker}, SenseNova-SI~\citep{cai-2025-sensenovasi}, ViLaSR~\citep{wu-2026-vilasr}, and MultiHopSpatial~\citep{lee-2026-multihopspatial}. The SpatialLadder-3B, SpatialThinker-3B/7B, SenseNova-SI-1.1-Qwen2.5-VL-3B/7B, SenseNova-SI-1.1-InternVL3-2B, and MultiHopSpatial-Qwen3-VL-4B-Instruct repositories declare Apache-2.0 licenses. The ViLaSR-7B repository does not explicitly state a checkpoint license in its model card or repository files. We use all checkpoints for research evaluation, and their redistribution remains subject to the applicable repository and upstream model terms.

\paragraph{Evaluation benchmarks.}
We evaluate on CV-Bench~\citep{tong-2024-cambrian}, BLINK~\citep{fu-2024-blink}, SpatialEval~\citep{wang-2024-spatialeval}, FRIEDA~\citep{pyo-2026-frieda}, and MultiHopSpatial~\citep{lee-2026-multihopspatial} using their publicly distributed evaluation resources. The CV-Bench and BLINK dataset repositories are licensed under Apache-2.0. MultiHopSpatial releases its question-answer annotations under Apache-2.0, while its COCO~\citep{lin-2014-coco} and PACO-Ego4D~\citep{ramanathan-2023-paco} images retain their original terms. We did not identify explicit dataset-wide license declarations for SpatialEval and FRIEDA. We use these resources for evaluation and do not assume redistribution rights beyond those granted by their respective providers.

\paragraph{Software and intended use.}
We use PostGIS\footnote{\url{https://postgis.net/}} for executing spatial queries over recorded scene geometries. PostGIS is distributed under the GNU General Public License. The generated supervision and trained models are intended for research and evaluation of spatial reasoning capabilities, rather than deployment in safety-critical applications. Redistribution of any third-party models, benchmark data, or modified software must comply with their respective licenses.

\section{Disclosure of AI Assistance}
\label{app:ai-disclosure}

We used LLMs to assist with manuscript
preparation, including language refinement,
structural revisions, and LaTeX formatting.
The authors reviewed and revised the generated
suggestions and retain responsibility for the
scientific content, experimental results, and
conclusions.

Separately, Qwen3-32B was used to paraphrase
canonical questions and execution-grounded
reasoning traces as part of the proposed data
generation pipeline (Appendix~\ref{app:method_paraphrasing}).
The model was not used to determine spatial
relations, intermediate states, or ground-truth
answers, which were derived through spatial
query execution.